\documentclass{article}[11pt]
\usepackage{graphicx} 
\usepackage{amsmath}
\usepackage{amssymb}
\usepackage{amsthm}
\usepackage{xurl}
\usepackage{mathtools}
\usepackage{natbib}
\usepackage{graphicx}
\usepackage{caption}
\usepackage{subcaption}
\usepackage{chngcntr}
\usepackage{apptools}
\usepackage{xcolor}
\definecolor{DarkGreen}{rgb}{0.1,0.5,0.1}
\definecolor{DarkRed}{rgb}{0.5,0.1,0.1}
\definecolor{DarkBlue}{rgb}{0.1,0.1,0.5}
\definecolor{HighlightOrange}{rgb}{1, 0.647, 0}
\definecolor{HighlightRed}{rgb}{0.8, 0.2, 0.2}
\usepackage{csquotes}
\usepackage{fancyhdr}
\usepackage{thmtools,thm-restate}
\usepackage{hyperref}
\hypersetup{
	colorlinks=true,       
	linkcolor=DarkBlue,    
	citecolor=DarkBlue,    
	filecolor=DarkBlue,    
	urlcolor=DarkBlue,     
	pdftitle={},
	pdfauthor={},
}
\usepackage[capitalize,noabbrev]{cleveref}
\usepackage{float}
\usepackage{parskip}
\usepackage{authblk}
\MakeOuterQuote{"}

\DeclarePairedDelimiter\ceil{\lceil}{\rceil}
\DeclarePairedDelimiter\floor{\lfloor}{\rfloor}

\theoremstyle{plain}
\newtheorem{theorem}{Theorem}

\newtheorem{lemma}{Lemma}

\newtheorem{assumption}{Assumption}
\AtAppendix{\counterwithin{lemma}{section}}
\AtAppendix{\counterwithin{theorem}{section}}
\AtAppendix{\counterwithin{prop}{section}}
\newtheorem{conjecture}{Conjecture}

\DeclareMathOperator*{\E}{\mathbb{E}}
\let\Pr\relax
\DeclareMathOperator*{\Pr}{\mathbb{P}}
\newcommand{\Acc}[1]{\mathrm{Acc}_{#1}}

\usepackage{mleftright}

\graphicspath{ {./images/} }

\definecolor{DarkRed}{rgb}{0.5,0.1,0.1}

\usepackage{ifthen}
\newboolean{icml}   
\setboolean{icml}{false}   

\newboolean{workshop}  
\setboolean{workshop}{false}   

\ifthenelse{\boolean{icml}}{\usepackage[accepted]{icml2024} \icmltitlerunning{Don't Label Twice:  Quantity Beats Quality when Comparing Binary Classifiers on a Budget}
}{\usepackage{fullpage}\setlength\parindent{0pt} \title{Don't Label Twice:  Quantity Beats Quality when Comparing Binary Classifiers on a Budget}}

\ifthenelse{\boolean{workshop}}{\usepackage{iclr2024_conference}}{}

\ifthenelse{\boolean{icml}}{}{
\author[1,2]{Florian E. Dorner\thanks{Corresponding author: florian.dorner@tuebingen.mpg.de}} 
\author[1,3]{Moritz Hardt}
\affil[1]{Max Planck Institute for Intelligent Systems, Tübingen}
\affil[2]{ETH Zürich}
\affil[3]{Tübingen AI Center}
}

\begin{document}

\ifthenelse{\boolean{icml}}{\twocolumn[
\icmltitle{Don't Label Twice:  Quantity Beats Quality when Comparing Binary Classifiers on a Budget}
\begin{icmlauthorlist}
\icmlauthor{Florian E. Dorner}{tue,ai,eth}
\icmlauthor{Moritz Hardt}{tue,ai}
\end{icmlauthorlist}
\icmlaffiliation{tue}{Max Planck Institute for Intelligent Systems, Tübingen}
\icmlaffiliation{ai}{Tübingen AI Center}
\icmlaffiliation{eth}{ETH Zürich}
\icmlcorrespondingauthor{Florian E. Dorner}{florian.dorner@tuebingen.mpg.de}
\vskip 0.3in
]
\printAffiliationsAndNotice{}}{\maketitle}

\begin{abstract}
\noindent We study how to best spend a budget of noisy labels to compare the accuracy of two binary classifiers. It is common practice to collect and aggregate multiple noisy labels for a given data point into a less noisy label via a majority vote. We prove a theorem that runs counter to conventional wisdom. If the goal is to identify the better of two classifiers, we show it’s best to spend the budget on collecting a single label for more samples. Our result follows from a non-trivial application of Cramér’s theorem, a staple in the theory of large deviations. We discuss the implications of our work for the design of machine learning benchmarks, where they overturn some time-honored recommendations. In addition, our results provide sample size bounds superior to what follows from Hoeffding’s bound.  
\end{abstract}
\section{Introduction}
Data annotators are the ``AI revolution's unsung heroes,'' \citet{gray2019ghost} argued. The labor of human annotators has powered a growing industry of machine learning datasets and benchmarks since the 1980s \citep{hardtrecht2022patterns}. Human labels are a precious, yet unreliable resource. Errors easily creep into data labor at scale. The designer of a benchmark has to cope with the reality of conflicting labels for the same data point. 

Many benchmarks follow a common strategy. Each data point in a sample gets noisy labels from multiple human annotators. The candidate labels then determine a single label via an aggregation function, such as a majority vote in the case of binary labels. For a sample of size~$n$ and a choice of~$m$ labels per data point, the cost of this design scales as $mn.$ Although ubiquitous, we prove that this strategy is wasteful for creating the test set. 

When the goal is to compare the population accuracy of binary classifiers, it is better to sample $mn$ data points and collect a single noisy label for each. The basis of our main result is a simple mathematical model that captures the essential question. Data points are independent and identically distributed. For each data point, we can request an odd number~$m\ge 1$ of binary labels, drawn independently from a distribution that picks the correct label with some probability strictly greater than chance. We then aggregate the $m$ labels into a single label using a majority vote. Fix two classifiers, one better than the other in terms of population accuracy by some positive margin. We have a budget $k$ to spend on labels. Given an annotator number $m$, we can create a labeled sample of size $n=k/m$. We pick the classifier with the higher empirical accuracy on this sample. How should we pick an annotator number $m$ so as to maximize the probability of picking the better classifier? Our main theorem provides the answer.
\begin{theorem}[Informal]
For a sufficiently large sample budget $k$, the probability of identifying the better of two binary classifiers is maximized at $m=1$ labels per data point.
\end{theorem}
As a rough intuition, the gains in label accuracy from aggregation are outweighed by the loss in sample size and information about classifier disagreements. Formally, our result follows from a non-trivial---and rather lengthy---reduction to Cramér’s theorem. Cramér’s theorem is fundamental to the theory of large deviations. It provides precise control over tail probabilities, based on the Legendre transform of the logarithmic moment generating function. The theorem is asymptotic with respect to $n$, complicating the application to our problem. However, standard concentration inequalities, such as Hoeffding’s bound, are insufficient for our purposes as they only provide upper bounds. 

Our theorem extends to the case where label errors are correlated with classifier errors, possibly even in a data dependent way. It only fails in the unusual cases where label noise systematically aligns in favor of the worse classifier, the effect of aggregation on label quality systematically aligns in favor of the better classifier, or the cost of unlabelled data is large. While we prove our main theorem for sufficiently large sample sizes, we conjecture that the statement holds for all $n\ge 1$. We have verified the conjecture numerically in a vast parameter sweep spanning more than four billion values. The numerical tool we created for verifying the conjecture also serves as an effective way to calculate tight sample size requirements for given parameters and is available at \href{https://labelnoise.is.tuebingen.mpg.de}{https://labelnoise.is.tuebingen.mpg.de}. 

Our result applies to the case of many classifiers via the union bound. Here, it gives an answer to the question how many classifiers we can reliably rank in a machine learning benchmark. This question is commonly answered in theory by combining the union bound with Hoeffding’s inequality. We demonstrate that our bound permits exponentially more comparisons than the standard argument for the same sample budget. Figure~\ref{fig:sample_size} 
illustrates the improvement.
\ifthenelse{\boolean{icml}}{

\begin{figure*}[ht]
    \begin{subfigure}[b]{0.5\textwidth}
         \centering
         \includegraphics[width=\textwidth]{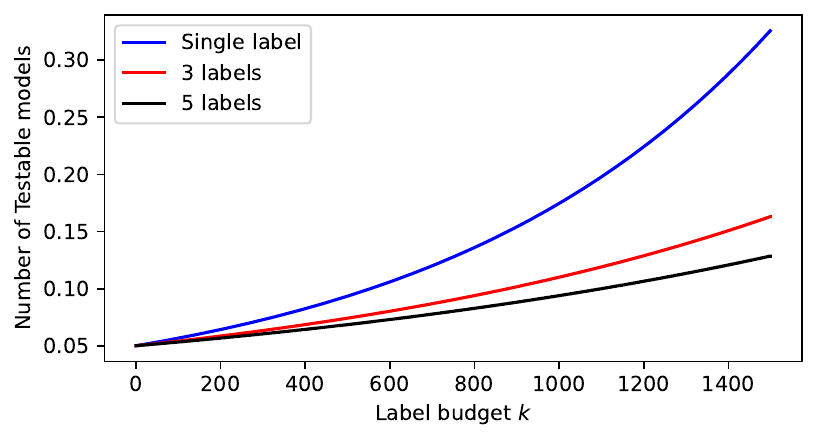}
         \caption{Hoeffding-based guarantees for $\delta=0.05$}
         \label{fig:sample_size_a}
     \end{subfigure}
     \hfill
     \begin{subfigure}[b]{0.5\textwidth}
         \centering
         \includegraphics[width=\textwidth]{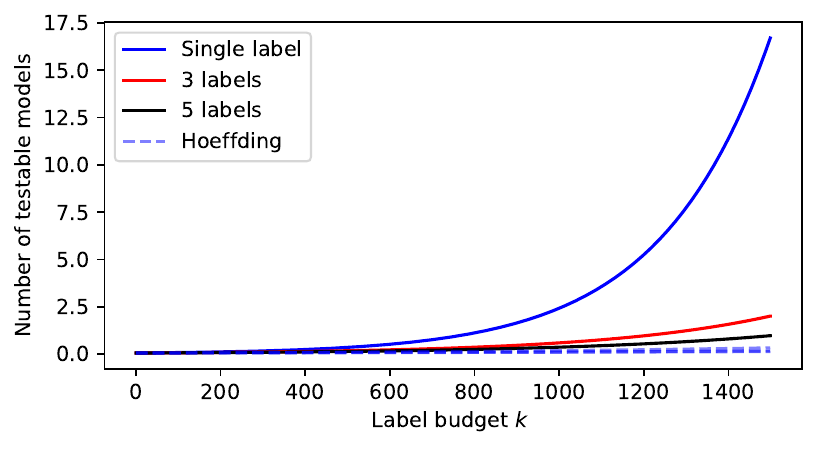}
         \caption{Cramér-based guarantees for $\delta=0.05$}
         \label{fig:sample_size_b}
     \end{subfigure}
\caption{Number of testable classifiers according to the Hoeffding (a) and Cramér-based (b) upper bounds on the error probability and a union bound (see Section \ref{sec:bench}) for accuracies $p=q=0.75$, margin $\epsilon=0.1$ and error tolerance $\delta=0.05$. Note the different $y$ axes. } 
         \label{fig:sample_size} 
\end{figure*}
}
{
\begin{figure*}[h]
     \centering
         \includegraphics[width=\textwidth]{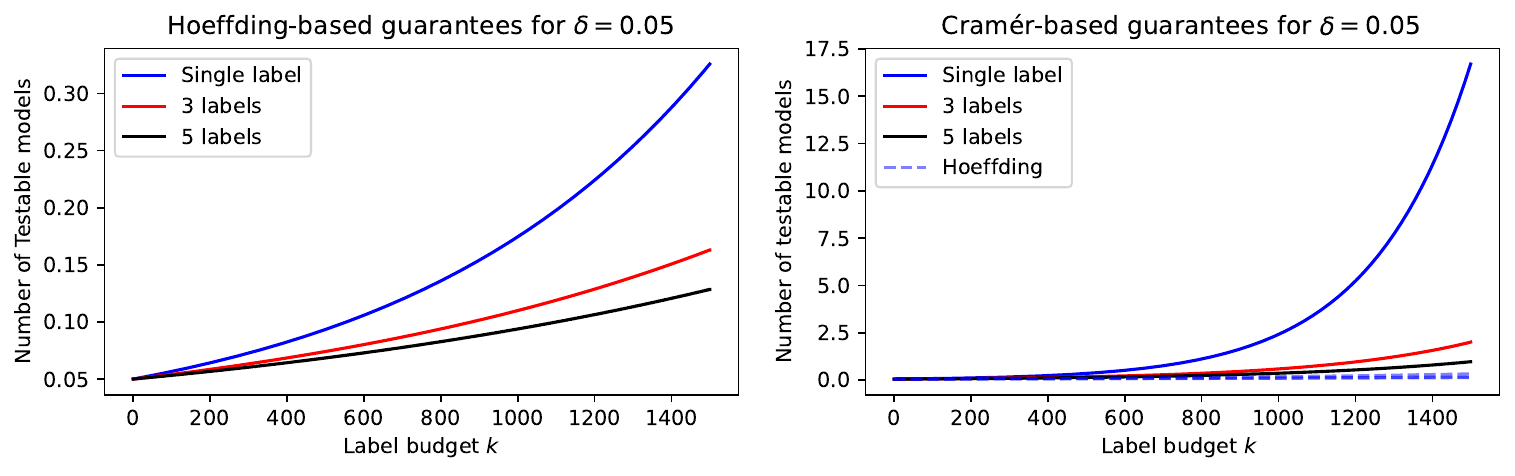}
         \caption{Number of testable classifiers according to the Hoeffding/Cramér based upper bounds on the error probability and a union bound for accuracies $p=q=0.75$, margin $\epsilon=0.1$ and error tolerance $\delta=0.05$. Note the different $y$ axes.} 
         \label{fig:sample_size} 
\end{figure*}}

There is a common belief that benchmark designers should invest in cleaning noisy labels through aggregation. Our result suggests a surprising departure. For the purpose of comparing and ranking binary classifiers, quantity beats quality. A single label per data point is optimal. 

\subsection{Related Work}\label{sec:related}
\paragraph{Label aggregation in dataset creation.}
Human-provided labels are at the heart of modern machine learning, both in industry \citep{gray2019ghost} and academic benchmarking. Many important datasets have been labeled by humans, with "gold standard" labels produced by aggregating multiple annotators' labels: In image recognition, labels for CIFAR-10 \citep{krizhevsky2009learning} were verified by the work's authors after being initially labeled by others, while labels in ImageNet \citep{russakovsky2015imagenet} are aggregated from multiple crowdworker annotations. Similarly, the target label for medical datasets is often established by a majority vote over expert annotators like sonographers \citep{tanno2019learning} or radiologists \citep{nguyen2022vindr}. In natural language processing, classic benchmarks like MSRP \citep{dolan2005automatically}, SST \citep{socher2013recursive}, SICK \citep{marelli2014semeval} and MNLI \citep{bowman2015large} all base labels on a per-instance majority vote after collecting multiple labels for each instance. More recently, label aggregation has been used to define test labels in Kaggle's Jigsaw Unintended Bias in Toxicity Classification challenge \citep{Jigsaw} and for evaluating the safety of LLama2 \citep{touvron2023llama}. Similarly, OpenAssistant \citep{kopf2023openassistant} aggregates users' rankings for the same list of model outputs into a "consensus opinion". \citet{recht2019imagenet} suggest to "employ a separate labeling process for the test set that relies on more costly expert annotations." In line with this, it is common to collect a larger amount of labels per instance for \textit{testing} than for training \citep{williams2017broad,dorner2022human,nguyen2022vindr} to increase label quality.
\paragraph{The impact of label aggregation on learning.}
While label aggregation is a common practice in dataset and benchmark creation, its impacts on training and evaluation are not fully understood: On the theoretical side, \citet{crammer2005learning} provide performance bounds that depend on the quality and size of training data and can be used to heuristically choose between data sources. \citet{wei2023aggregate} analyze whether duplicate labels for the same data point should be aggregated or treated independently for empirical risk minimization and find the latter to perform better if disagreement is common. Similarly \cite{cheng2022many}, find that maximum-likelihood estimation based on available duplicate labels outperforms majority voting for well-specified models. However, the authors also warn that model mis-specification can cause disaggregated MLE to break down, while majority voting is more robust. Along similar lines, when optimizing surrogate losses, majority voting can provide consistency in cases where directly learning from disaggregated data is impossible \cite{cheng2025some}. Empirically \citet{sheng2008get} show that for certain decision tree learners, a large number of noisy labels per instance $x$ beats single labels for more data points when labels are very noisy. On the other hand, \citet{chen2021clean} provide empirical evidence that for realistic label noise, the opposite is true for finetuning modern language models. In line with that, \citet{lin2014re} show that the benefits of relabeling can depend both on the problem domain and hyperparameters of the learning algorithm. In contrast to these works, our work focuses on comparing already learnt classifiers, not classifier training. 
\paragraph{Annotator disagreement as a feature.}
\citet{aroyo2013crowd} argue that due to the lack of objective ground truth for many tasks, taking annotator disagreement into account is essential. The authors suggest to use non-binary labels that encompass disagreement. \citet{ramponi2022dh} and \citet{sandri2023don} use predicting annotator disagreement as an auxiliary task for detecting offensive language, while \citet{cheplygina2018crowd} show that annotator disagreement itself can be an informative feature in medical image analysis. Meanwhile, \citet{tanno2019learning} and \citet{davani2022dealing} suggest to predict individual annotators' responses. This approach, combined with focusing on annotators relevant for a given context, is also used to mitigate majoritarian biases caused by aggregation \citep{gordon2022jury,fleisig2023fair}. As these approaches require annotator-level labels, it is often recommended for dataset creators to release these rather than already aggregated labels \cite{prabhakaran2021releasing,denton2021whose}. Our work is orthogonal: We focus on cases where the target label is agreed upon to be given by a (fictitious) majority vote over the whole crowdworker population. In this setting, we demonstrate that collecting and aggregating multiple labels per data point is \textit{statistically} suboptimal in terms of identifying the better of two classifiers. 
\paragraph{The theory of benchmarking.}
Benchmarking plays an important role in machine learning, but is rarely studied. An exception is work on \textit{adaptive overfitting}: For the test error to estimate the population risk without bias, models have to be trained without knowledge about the test set, which is rarely true for real benchmarks. To see whether this causes problems in practice, \citet{recht2019imagenet} recreated the ImageNet test set based on the original procedure. They find that classifier accuracy on the new test set is lower, but strongly correlates with the original accuracy such that model rankings are remarkably stable. \citet{mania2020classifier} theoretically explain these observations based on correlations between classifiers. Lastly, \citet{blum2015ladder} show that the impacts of adaptive overfitting can be reduced by only revealing a classifier's test accuracy if it is substantially better than the previous best.


\section{Formal Setup}\label{sec:base} 
Let $\mathcal{D}$ be a distribution of data points $x$ with binary correct labels $y_{\mathit{True}}\mleft(x\mright) \in \{0,1\}$. For a binary classifier $c$, we define the population risk as the expected frequency of classification errors \[\mathcal{R}\mleft(c\mright)\coloneqq \E_{x\sim \mathcal{D}} [\mathbb{I}\mleft(y_{\mathit{True}}\mleft(x\mright) \neq c\mleft(x\mright)\mright)],\] where $\mathbb{I}$ denotes the indicator function. We consider two arbitrary classifiers $c_b$ and $c_w$ (where $b$ stands for "better" and $w$ for "worse"), such that \[1-p= \mathcal{R}\mleft(c_w\mright) > \mathcal{R}\mleft(c_b\mright) = 1-p-\epsilon \] for accuracy $p \in [0.5,1]$ and margin $\epsilon\in (0,1-p]$. We want to use a limited labeling budget $k$ to create a test set $T$ on which test accuracy is likely to be higher for the better classifier, without using any information about the two specific classifiers at hand. We assume that test sets are created using the following sampling procedure: Independently (with replacement) sample a dataset $D$ of $n$ data points $x\sim \mathcal{D}$. Then, for each $x\in D$, sample  $m=\frac{k}{n}$ labelers $l$ from a population of crowdworkers $\mathcal{D}_{crowd}$, again independently and with replacement, and have each of them provide a label $y_l\mleft(x\mright)$ for $x$.  For a given data point $x$, we then set the test label $y_{\mathit{Test}}\mleft(x\mright)$ equal to the majority of the labels $y_l\mleft(x\mright)$. The main question tackled in this work is then, how to allocate the label budget $k$ between $n$ and $m$ in order to have the best chance of correctly identifying the better classifier $c_b$ using the constructed test set. We will particularly focus on comparing the case of $m=1$ to $m>1$, as we find strong evidence that $m=1$ is optimal in most cases. 

For a fixed data point $x$, we set $q\mleft(x\mright)\in (0.5,1]$ to the probability that a crowdworker label $y_l(x)$ is correct, marginalized over $l$, i.e. $q(x) \coloneqq \Pr_{l}\mleft(y_l\mleft(x\mright) = y_{\mathit{True}}\mleft(x\mright)\mright)$. Similarly, $q$ denotes the same probability marginalized over both $x$ and $l$: $q\coloneqq \Pr_{x,l}\mleft(y_l\mleft(x\mright) = y_{\mathit{True}}\mleft(x\mright)\mright)$. We note, that in this setup, the case of collecting $m$ labels $y_l\mleft(x\mright)$ for a given $x$ with correctness probability $q\mleft(x\mright)$ yields the same distribution of labels as collecting a single label with correctness probability $q'\mleft(x\mright)=M_m\mleft(q\mleft(x\mright)\mright)$, 
where 
\[
M_{m}\mleft(q\mright) \coloneqq \Pr(\text{Majority of $m$ independent voters correct})
\]
under the assumption that each voter is correct with probability $q$.
To compare the two classifiers $c_b$ and $c_w$ on our test set, we define the gap indicator $G$
\[
    G \coloneqq 
    \begin{cases}
    1: &c_b\mleft(x\mright) = y_{\mathit{Test}\mleft(x\mright)} \neq c_w\mleft(x\mright)   \\
    -1: &c_w\mleft(x\mright) = y_{\mathit{Test}\mleft(x\mright)} \neq c_b\mleft(x\mright)    \\
    0: &c_w\mleft(x\mright) = c_b\mleft(x\mright) 
    \end{cases}\,,
\]
where $x$ and $y_{\mathit{Test}}$ are sampled as described above. The gap indicator $G$ describes the unnormalized accuracy gap between the classifiers $c_b$ and $c_w$ on the test set, as we can express 
\[ \frac{1}{n}\sum^{n}_{i=1} G_i  =  \Acc{\mathit{Test}}\mleft(c_b\mright) -  \Acc{\mathit{Test}}\mleft(c_w\mright)
\] for independent copies $G_i$ of $G$, where $ \Acc{\mathit{Test}}\mleft(c\mright) \coloneqq \frac{1}{n}\sum^{n}_{i=1} \mathbb{I}\mleft(y_{\mathit{Test}}\mleft(x\mright) = c\mleft(x\mright)\mright)$. In particular $ \sum^{n}_{i=1} G_i $ is positive if and only if our test set correctly  identifies the better classifier $c_b$. 
\section{Parameterizing the Gap Indicator 
}\label{sec:G_Params}
We begin by considering the fully independent case in which the error events $c_w\mleft(x\mright) \neq y_{\mathit{True}}\mleft(x\mright)$, $c_b\mleft(x\mright) \neq y_{\mathit{True}}\mleft(x\mright)$ are independent of each other and the label accuracy $q(x)$. We will treat the gap indicator $G$ as a function of $q$ and assume homogeneous label errors over $x$, i.e. $q\mleft(x\mright)=q$. This assumption allows us to use the equivalence of a single labeler with accuracy $M_m\mleft(q\mleft(x\mright)\mright)$ and $m$ labelers with accuracies $q\mleft(x\mright)$ each, to compare $G\mleft(q\mright)$ and $G\mleft(M_m\mleft(q\mright)\mright)$ rather than explicitly parameterizing $G$ by $m$. We note that this assumption yields the best case for the $m-$label approach: If the label accuracy $q\mleft(x\mright)$ depends strongly on $x$, majority voting might not actually yield noticeable benefits in terms of label accuracy. As an extreme example, if $q\mleft(x\mright)$ only takes on values in $\{0,1\}$, majority voting has no benefits at all. Formally, Jensen's inequality and the well-known concavity of the majority vote in $M_m\mleft(z\mright)$ in $z$ for $z\in (0.5,1]$ \citep{boland1989modelling} imply $\E_x[M_m\mleft(q\mleft(x\mright)\mright)]\leq M_m\mleft(\E_x[q\mleft(x\mright)]\mright)=M_m\mleft(q\mright)$. This means that $M_m\mleft(q\mright)$ can only overestimate label quality for the $m$-label case. The following proposition provides a precise parametric characterization for $G$ with $m=1$ in that case, and is proven in Appendix~\ref{app:params}.
\begin{restatable}{prop}{propindep}
Assuming mutually independent classifier and labeler errors, $G$ can be written as follows:
\[
    G\mleft(q,p,\epsilon\mright) = 
    \begin{cases}
    1 &\text{ w.p. } q\epsilon + \mleft(1-p-\epsilon\mright)p  \\
    -1 &\text{ w.p. } \mleft(1-q\mright)\epsilon  + \mleft(1-p-\epsilon\mright)p  \\
    0 &\text{ w.p. }  p\mleft(p+\epsilon\mright) + \mleft(1-p-\epsilon\mright)\mleft(1-p\mright) 
    \end{cases},
\]
for label accuracy $q$, classifier accuracy $p$ and margin $\epsilon$. 
\end{restatable}
The expectation of the random variable $G\mleft(q,p,\epsilon\mright)$ thus equals $\mleft(2q-1\mright)\epsilon$, which is positive as $q>0.5$. 
 
We are now interested in whether using an $m-$majority vote of the noisy crowdworker labels provides more information about which of the two classifiers is better than using $m$ times as many data points with a single label each. More precisely, we would like to find out whether the better classifier is more likely to win with a single label and more data points, or with aggregated labels. In technical terms, 
we thus want to know whether \[\Pr\mleft(\sum_{i=1}^{mn} G_i\mleft(q,p,\epsilon\mright)>0\mright)>\Pr\mleft(\sum_{i=1}^n G_i\mleft(M_{m}\mleft(q\mright),p,\epsilon\mright)>0\mright)\] for independent copies $G_i$ of $G$. For small $n$ and $m$ we can calculate the exact probabilities of identifying the better classifier $c_b$ using test sets with different levels of label accuracy $q$. According to a large scale grid search over the possible values of $p,q$ and $\epsilon$ that evaluated nearly five billion configurations, detailed in Appendix \ref{sec:experiments}, using $m=1$ labels is consistently the best approach. Figure \ref{fig:exact} demonstrates this, showing the exact probabilities for fixed accuracy $p=0.8$, margin $\epsilon=0.01$ and varying values of the label accuracy $q$ and label budget $k$. 

\ifthenelse{\boolean{icml}}{

\begin{figure*}[ht]
    \begin{subfigure}[b]{0.5\textwidth}
         \centering
         \includegraphics[width=\textwidth]{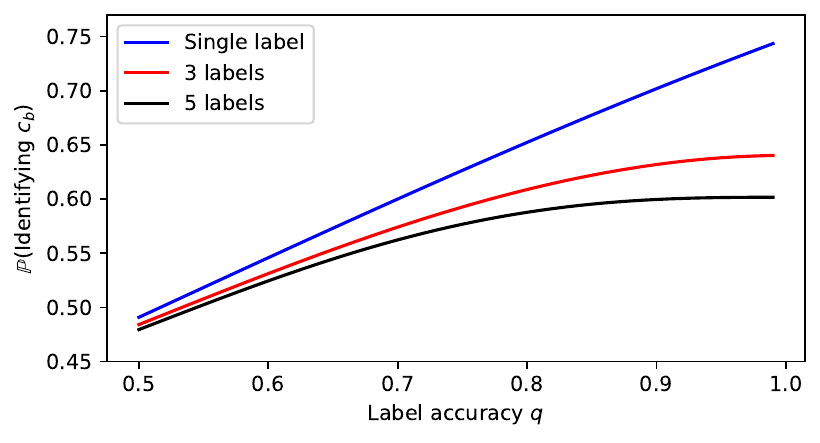}
         \caption{Exact probabilities for $p=0.8$, $\epsilon=0.01$, $k=1500$}
         \label{fig:exact_a}
     \end{subfigure}
     \hfill
     \begin{subfigure}[b]{0.5\textwidth}
         \centering
         \includegraphics[width=\textwidth]{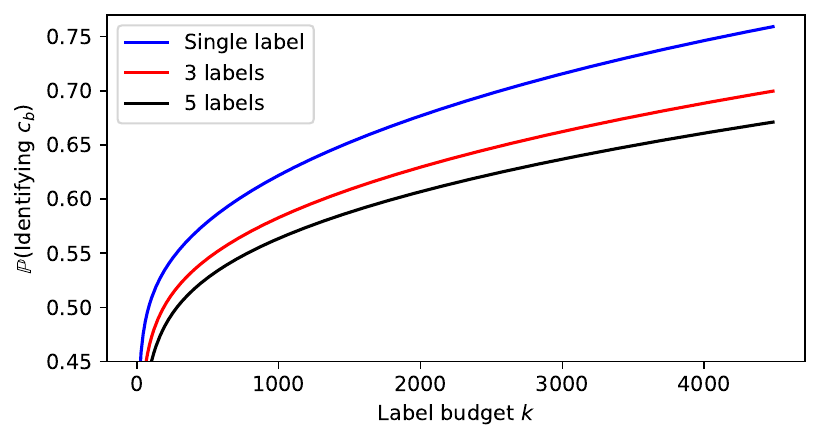}
         \caption{Exact probabilities for $p=0.8$, $\epsilon=0.01$, $q=0.8$}
         \label{fig:exact_b}
     \end{subfigure}
    \caption{Probability of identifying $c_b$ for accuracy $p=0.8$, margin $\epsilon=0.01$, budget $k=1500$ (a), label accuracy $q=0.8$ (b).} 
    \label{fig:exact}
\end{figure*}
}
{
\begin{figure*}[ht]
     \centering
         \includegraphics[width=\textwidth]{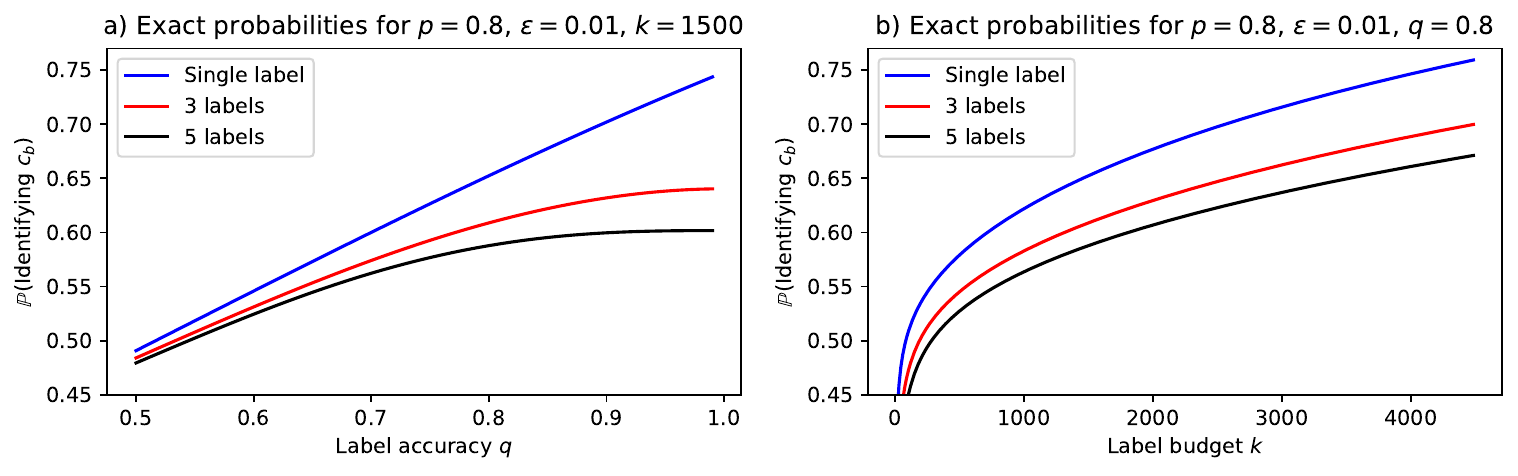}
         \caption{Probability of identifying $c_b$ for accuracy $p=0.8$, margin $\epsilon=0.01$, budget $k=1500$ (a), label accuracy $q=0.8$ (b).} 
         \label{fig:exact}
\end{figure*}
}

\subsection{Hoeffding Bounds}\label{sec:hoeffding}
Hoeffding's inequality yields the following lemma proven in Appendix \ref{app:hoeffding} that allows us to lower bound the probability that $c_b$ beats $c_w$ on a test set: 
\begin{restatable}{lemma}{lemmahoeffding}\label{lemma:hoeff}
For independent copies $X_i$ of any random variable $X$ with $\E[X]>0$ and values in $[-1,1]$, we can bound \[\Pr\mleft(\sum^n_{i=1} X_i \leq 0\mright) \leq  e^{ \frac{- n \E[X]^2 }{2}}\eqqcolon B(X,n).\] 
\end{restatable}
We will use this lemma to gain some initial intuition about the quality of test sets constructed with $m=1$, compared to $m>1$ labelers per data point $x$. Specifically, we get a higher lower bound for $\Pr\mleft(\sum_{i=1}^{mn} G_i\mleft(q,p,\epsilon\mright)>0\mright)$ than for $\Pr\mleft(\sum_{i=1}^n G_i\mleft(M_{m}\mleft(q\mright),p,\epsilon\mright)>0\mright)$, 
whenever \begin{equation}\label{hoeffding_abstract}
nm \E[G_i\mleft(q,p,\epsilon\mright)]^2  > n \E[G_i\mleft(M_{m}\mleft(q\mright),p,\epsilon\mright)]^2. 
\end{equation} Informally, equation \eqref{hoeffding_abstract} states that the gains in terms of squared expectation from aggregating multiple labels do not outweigh the simple factor $m$ achieved by labeling multiple data points. It is equivalent to  \begin{equation}\label{hoeffding}
\sqrt{m}   >  \frac{\E[G_i\mleft(M_{m}\mleft(q\mright),p,\epsilon\mright)] }{\E[G_i\mleft(q,p,\epsilon\mright)]} =  \frac{2M_m\mleft(q\mright)-1 }{2q-1}. 
\end{equation} 
For $m=3$, this becomes \[ \sqrt{3}   >  \frac{6q^2 - 4q^3 -1 }{2q-1} = 1 + 2 \mleft(1-q\mright)\mleft(q\mright),\] with the right side maximized at $q=0.5,$ with a value of $1.5<1.73 \approx \sqrt{3}$, such that \eqref{hoeffding} is true for all $q\in (0.5,1]$. In Appendix \ref{app:hoeffding}, we prove that $B\mleft(G\mleft(q,p,\epsilon\mright),mn\mright) < B\mleft(G\mleft(M_{m}\mleft(q\mright),p,\epsilon\mright),n\mright)$ holds for any $m>1$. 

\subsection{Correlated Classifiers}\label{sec:corr_label}
The previous Sections assumed both classifiers and the labels to be independent, which is unlikely in practice, as certain examples might be more difficult than others. In this Section, we relax this assumption by modelling the worse classifier $c_w$ to be correct with probability $p_w \in [0.5,1]$. Then, the better classifier $c_b$ is  correct with probability $p_b^0 \in [0.5,1]$ conditional on $c_w$ being incorrect on a given datapoint $x$ and $p_b^1\in [0.5,1]$ conditional on $c_w$ being correct. The assumption that $c_b$ has lower risk than $c_w$ implies $ \mleft(1-p_w\mright) p_b^0 + p_w p_b^1  > p_w$ or equivalently $ \mleft(1-p_w\mright) p_b^0 + p_w \mleft(p_b^1-1\mright)  > 0$. We also model correlations between the two classifiers and the labels by denoting $q_b\in (0.5,1]$ as the probability that the label is correct, conditional on the event $E_b$ that $c_b$ is correct and $c_w$ is incorrect, and $q_w\in (0.5,1]$ as the probability that the label is correct in the case that $c_b$ is incorrect and $c_w$ is correct, termed $E_w$.\footnote{As data points $x$ for which both agree do not influence the gap indicator, label accuracy can be arbitrary for such points, i.e. $x$ that are neither in $E_b$ nor in $E_w$.}

\begin{restatable}{prop}{propdeplabel}
Assuming correlated classifiers and labels with the above parameterization, we have:
\[
    G\mleft(q,p\mright) = 
    \begin{cases} 
    1 &\text{w.p. } q_b \mleft(1-p_w\mright) p_b^0 \ifthenelse{\boolean{icml}}{\nonumber \\&}{}+  \mleft(1-q_w\mright) p_w \mleft(1-p_b^1\mright)\\
    -1 &\text{w.p. }  \mleft(1-q_b\mright) \mleft(1-p_w\mright) p_b^0 \ifthenelse{\boolean{icml}}{\nonumber \\&}{}+  q_w p_w \mleft(1-p_b^1\mright) \\
    0 &\text{else }  
    \end{cases}.
\]
\end{restatable}
Then, the expectation of the gap indicator $G$ equals
\begin{align*}
\mleft(2q_b-1\mright) \mleft(1-p_w\mright) p_b^0 - \mleft(2q_w-1\mright)  p_w \mleft(1-p_b^1\mright), 
\end{align*} which is larger than zero if and only if \[  \mleft(2q_b-1\mright) > \mleft(2q_w-1\mright)  \frac{p_w\mleft(1-p_b^1\mright)}{\mleft(1-p_w\mright) p_b^0}.
\]
The factor  $\frac{p_w\mleft(1-p_b^1\mright)}{\mleft(1-p_w\mright) p_b^0}$ is smaller than one, as long as $\mleft(1-p_w\mright) p_b^0 - p_w \mleft(1-p_b^1\mright) >0$, which is true as $c_b$ has lower risk than $c_w$. This means that $G$ is guaranteed to have positive expectation, whenever $q_b\geq q_w$. This essentially ensures that $c_w$ is not overfit to the label noise more than $c_b$. We assume this to be true: 
\begin{assumption}\label{ass:qs} 
No biased label accuracy: $q_b\geq q_w$.
\end{assumption}
An alternative interpretation of assumption \ref{ass:qs} is, that the examples in $E_w$, for which the better classifier is incorrect, are more “difficult” than the ones in $E_b$. For example, consider $c_b$ correct on all but for the top 10\% most difficult examples, and $c_w$ random. Then $E_w$ is a subset of the top 10\% most difficult examples, such that annotators make more errors and $q_w$ is low. At the same time, $E_b$ only contains examples in the bottom 90\%, such that $q_b$ is large.

If assumption \ref{ass:qs} does not hold, for example because $c_w$ was trained on parts of the test set, the expectation of $G$ can become negative, such that $\Pr\mleft(\sum_i^n G_i >0\mright)$ converges to zero. In these cases, narrowing the gap between $q_b$ and $q_w$ by aggregating labels ($M_m\mleft(q_w\mright)\approx M_m\mleft(q_b\mright)$ for large $m$) can have large benefits by causing the expectation to become positive, thus flipping the limit of $\Pr\mleft(\sum_i^n G_i >0\mright)$ from zero to one.

For the $m-$label case, we again focus on (conditionally) homogeneous label errors over $x$, i.e. $q\mleft(x\mright)=q_b$ when $x\in E_b$ and $q\mleft(x\mright)=q_w$ when $x\in E_w$, so that we can replace $G\mleft(q_b,q_w\mright)$ by $G\mleft(M_m\mleft(q_b\mright),M_m\mleft(q_w\mright)\mright)$ rather than explicitly parameterizing $G$ by $m$. Note that in this case, homogeneity in the label accuracy $q\mleft(x\mright)$ is not necessarily the best case for $m>1$ any more: Heterogeneity lowering the label accuracy of the majority vote can be beneficial as long it is restricted to $E_w$, where $c_b$ is incorrect. We assume that heterogeneity does not disproportionately harm label accuracy when the better classifier is incorrect:

\begin{assumption}\label{ass:jensen} No biased heterogeneity:
    \begin{align}
    \ifthenelse{\boolean{icml}}{&}{}
    \frac{\mleft(1-p_w\mright)p_b^0}{p_w \mleft(1-p_b^1\mright)}\mleft(M_m\mleft(q_b\mright)-\E_{x}[M_m\mleft(q\mleft(x\mright)\mright)|E_b]\mright) \ifthenelse{\boolean{icml}}{\nonumber \\&}{} \geq   M_m\mleft(q_w\mright)-\E_{x}[M_m\mleft(q\mleft(x\mright)\mright)|E_w].\end{align}
\end{assumption}
The $\frac{\mleft(1-p_w\mright)p_b^0}{p_w \mleft(1-p_b^1\mright)}$ factor is larger than one as $c_b$ is more accurate than $c_w$. Because $q_b>q_w$ and $M_3\mleft(x\mright)$ is more concave for larger $x>0.5$, this means that for $m=3$ assumption~\ref{ass:jensen} is expected to hold whenever there are similar levels of heterogeneity conditional on the events $E_b$ and $E_w$. For simplicity of notation, we will sometimes use $p$ as a shorthand for $p_w,p_b^0,p_b^1$, $q$ as a shorthand for $q_b,q_w$ and $M_m\mleft(q\mright)$ as a shorthand for $M_m\mleft(q_b\mright),M_m\mleft(q_w\mright)$, again obtaining $B\mleft(G\mleft(q,p\mright),nm\mright) > B\mleft( G\mleft(M_{m}\mleft(q\mright),p\mright),n\mright)$ for any $m>1$ under assumption \ref{ass:qs}, as proven in Appendix \ref{app:hoeffding}.

\subsection{Application to Benchmarking}\label{sec:bench}
The different bounds on the error probabilities for a single vs $m$ labels are straightforward to extend to benchmarking, where we compare multiple classifiers: Formally, we consider a classifier $c_b$ with risk $\mathcal{R}\mleft(c_b\mright)=1-p-\epsilon$ that is better than $k$ other classifiers $c_i, i\leq k$ with (larger) risk $\mathcal{R}\mleft(c_i\mright)\geq 1-p$. A test set is a good benchmark, if $c_b$ has the highest test accuracy with high probability. We can bound the probability that the benchmark fails to identify the best classifier $c_b$ using a standard union bound argument: \begin{align*}   
& \Pr\mleft(\Acc{\mathit{Test}}\mleft(c_b\mright)\le \max_{i\leq k} \Acc{\mathit{Test}}\mleft(c_i\mright)\mright) 
\ifthenelse{\boolean{icml}}{\\&}{}
\leq \sum_{i\leq k} \Pr\mleft(\Acc{\mathit{Test}}\mleft(c_b\mright) \le \Acc{\mathit{Test}}\mleft(c_i\mright)\mright).
\end{align*}
Now if $\Pr\mleft(\Acc{\mathit{Test}}\mleft(c_b\mright)\le \Acc{\mathit{Test}}\mleft(c_i\mright)\mright)\leq e^{-d n\epsilon^2}$ for some $d>0$ and all $i\leq k$ as suggested by the Hoeffding bounds from the last Section, we get \[\delta \coloneqq \Pr\mleft(\Acc{\mathit{Test}}\mleft(c_b\mright)\le\max_{i\leq k} \Acc{\mathit{Test}}\mleft(c_i\mright)\mright) \leq k e^{-dn \epsilon^2 }.\] If we want to bound the probability of not identifying the best classifier $c_b$ to a fixed $\delta>0$, we can thus test at most $k=e^{d  n\epsilon^2} \delta$ different classifiers. Correspondingly under the assumptions from before, moving from an $e^{-d_1  n\epsilon^2}$ to an $e^{-d_2  n\epsilon^2}$ bound for $d_2>d_1$
by not collecting multiple labels per data point allows us to benchmark $e^{\mleft(d_2-d_1\mright)  n\epsilon^2}$ times as many classifiers while guaranteeing a given bound on the error probability $\delta$. 

This exponential improvement is illustrated in Figure \ref{fig:sample_size},
which also illustrates the lack of tightness of Hoeffding bounds in our setting, when compared to the bounds provided by Cramér's Theorem discussed in the next Section: For a label budget of $k=1500$, Cramér guarantees the testability of more than $17$ models in the single label case, while Hoeffding is too loose to provide a guarantee for two models at error tolerance $\delta=0.05$.

\section{Proof of the Main Theorem}\label{sec:ld_main}
The results proven above are suggestive, but do not prove that a single label is optimal. This is because we compare lower bounds that could have systematically different levels of tightness for the single label compared to the $m-$label case.  As a large test set not correctly identifying the better classifier is a tail event, we use tools from the theory on large deviations, more specifically Cramér's Theorem to provide a proof for sufficiently large values of~$n$.

\begin{restatable}{theorem_cramer}{cramer} (Adapted from \citep{klenke2013probability}) Let $X_i$ be iid real random variables for $i \in \mathbb{N}$ such that \[\Lambda\mleft(t\mright)\coloneqq\log \E[e^{tX_{1}}]<\infty\] for all $t \in \mathbb{R}$. 
Define the Legendre transform \[\Lambda^*\mleft(x\mright)\coloneqq\sup_t \mleft(tx-\Lambda\mleft(t\mright)\mright).\] Then for all $z\in \mathbb{R}$ such that $z>\E[X_1]$, we have \[\lim_{n\to\infty}\frac{1}{n} \log \Pr\mleft(S_n=\sum_{i=1}^n X_i\geq zn\mright) = - \Lambda^*\mleft(z\mright),\] where the limit is an upper bound for all $n$.  
\end{restatable}
We apply the theorem to the random variables $X = -G\mleft(M_m\mleft(q\mright),p,\epsilon\mright)$ and $X' = -\sum_{i=1}^m G_i\mleft(q,p,\epsilon\mright)$ at $z=0$, which is possible, as both $-X$ and $-X'$ have positive expectation, such that $z=0>\E[X]$. This yields limits 
\begin{align*} \lim_{n\to\infty} \frac{1}{n} \log \mleft( \Pr\mleft(\sum^n_i G_i\mleft(M_m\mleft(q\mright),p,\epsilon\mright)\leq 0\mright)\mright)  =  -\Lambda^*_X\mleft(0\mright) \end{align*} and \[ \lim_{n\to\infty} \frac{1}{n} \log \mleft( \Pr\mleft(\sum^{mn}_i G_i\mleft(q,p,\epsilon\mright)\leq 0\mright)\mright) = -\Lambda^*_{X'}\mleft(0\mright)\] respectively. Because $\Pr(X\leq 0) = 1 - \Pr(X>0)$ for any random variable $X$, we can conclude that  \[\Pr\mleft(\sum^{mn}_i G_i\mleft(q,p,\epsilon\mright)>0\mright) > \Pr\mleft(\sum^n_i G_i\mleft(M_m\mleft(q\mright),p,\epsilon\mright)>0\mright)\] will be true for sufficiently large $n$ as long as $-\Lambda^*_X\mleft(0\mright)>-\Lambda^*_{X'}\mleft(0\mright)$.
Figure \ref{fig:cramer} a) illustrates the convergence implied by Cramér's theorem for a fixed set of parameters. As the Cramér rates are upper bounds, we can conclude that the single label approach is better, as soon as the absolute gap between the Cramér rates exceeds the maximum of the approximation errors (here around $k=1800$). Meanwhile, Figure \ref{fig:cramer} b) shows the tightness of Cramér's bound compared to Hoeffding's bound. While both are very close when labels are random ($q=0.5$),  Cramér's bound becomes a lot smaller when labels are accurate. 

\ifthenelse{\boolean{icml}}{

\begin{figure*}[ht]
    \begin{subfigure}[b]{0.5\textwidth}
         \centering
         \includegraphics[width=\textwidth]{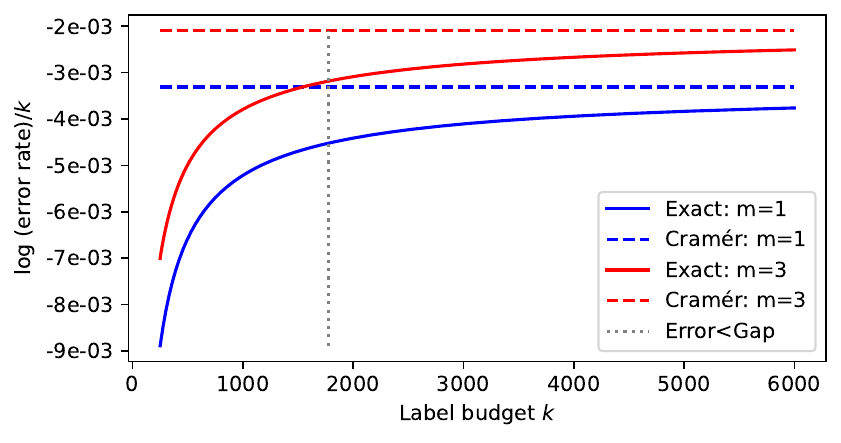}
         \caption{Error rates for $q=0.75$, $p=0.7$, $\epsilon=0.1$}
         \label{fig:cramer_a}
     \end{subfigure}
     \hfill
     \begin{subfigure}[b]{0.5\textwidth}
         \centering
         \includegraphics[width=\textwidth]{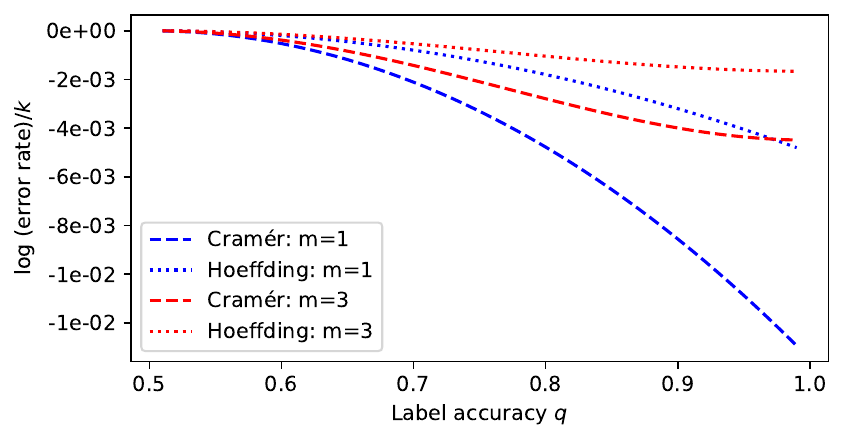}
         \caption{Error bounds for $k=2000$, $p=0.7$, $\epsilon=0.1$}
         \label{fig:cramer_b}
     \end{subfigure}
         \caption{a): Convergence of normalized log error rates to the values implied by Cramér's Theorem for label accuracy $q=0.75$, classifier accuracy $p=0.7$, margin $\epsilon=0.1$ and $m\in\{1,3 \}$. b): Upper bounds on normalized log error rate for Cramér's bound compared to Hoeffding's bound.} 
         \label{fig:cramer}
\end{figure*}
}{
\begin{figure*}[ht]
     \centering
     \includegraphics[width=\linewidth]{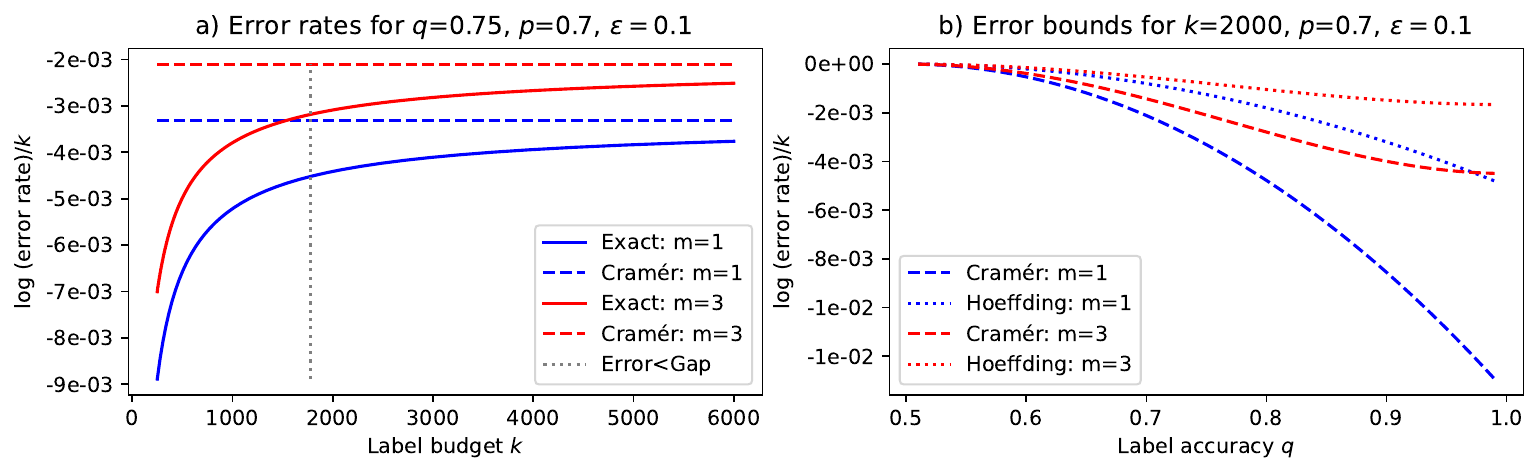}
         \caption{a): Convergence of normalized log error rates to the values implied by Cramér's Theorem for label accuracy $q=0.75$, classifier accuracy $p=0.7$, margin $\epsilon=0.1$ and $m\in\{1,3 \}$. b): Upper bounds on normalized log error rate for Cramér's bound compared to Hoeffding's bound.} 
         \label{fig:cramer}
\end{figure*}}

To prove $-\Lambda^*_X\mleft(0\mright)>-\Lambda^*_{X'}\mleft(0\mright)$, we make use of the simple ternary structure of $G$ and the following lemma characterising $-\Lambda^*_X\mleft(0\mright)$ for ternary random variables:
\begin{restatable}{lemma}{ternary}\label{lemma:explicit_lambda}
For $X$ ternary with $\Pr\mleft(X=1\mright) = x$, $\Pr\mleft(X=-1\mright) = y$, and $\Pr\mleft(X=0\mright) = z$, \[-\Lambda_X^*\mleft(0\mright)=\log{\mleft(2\sqrt{xy}+z\mright)}.\] For sums of independent copies $X_i$ of ternary variables $S_n=\sum_{i=1}^n X_i,$ \[-\Lambda_{S_n}^*\mleft(0\mright)=n\log{\mleft(2\sqrt{xy}+z\mright)}.\]
\end{restatable}

Lemma \ref{lemma:explicit_lambda} can then be used to prove our main theorem: 
\begin{restatable}{theorem}{main} \label{thm:asym_optim}
    For $G$ as defined as in Section \ref{sec:corr_label}, $m>1$ and $q_b,q_w,p_w,p_b^0,p_b^1$ fixed such that assumption \ref{ass:qs} holds, there exist an $N\in \mathbb{N}$ such that for $n>N$ \[\Pr\mleft(\sum^n_i G_i\mleft(M_m\mleft(q\mright),p\mright)>0\mright)<\Pr\mleft(\sum^{mn}_i G_i\mleft(q,p\mright)>0\mright).\] 
    Under assumption \ref{ass:jensen}, this implies that the single label strategy outperforms the $m-$label strategy for these $n$.
    \end{restatable}

In other words, under the assumptions from Section \ref{sec:corr_label} on the joint distribution of the labels and classifiers and sufficiently large label budgets $mn$, it is always better to collect a single label for $mn$ data points rather than $m$ labels for $n$ data points, when it comes to  classifier comparison. We note that the assumptions in Section \ref{sec:corr_label} generalize those from Section \ref{sec:base} by taking into account correlations between classifiers and labels, such that Theorem \ref{thm:asym_optim} also holds for the independent case discussed in Section \ref{sec:base}.  

\ifthenelse{\boolean{workshop}}{We sketch the proof on a high level:}{We begin by sketching the proof on a high level:} First, we observe that  $-m \Lambda^*_X\mleft(0\mright) = -\Lambda^*_{X'}\mleft(0\mright)$ whenever $q=q_b=q_w=0.5$, as $M_m\mleft(0.5\mright)=0.5$. As both of these terms have to be negative, this establishes $-\Lambda^*_X\mleft(0\mright)> -m \Lambda^*_X\mleft(0\mright) = -\Lambda^*_{X'}\mleft(0\mright)$. We then show that in the setting of Section \ref{sec:base}, the derivative of $-\Lambda^*_X\mleft(0\mright) + \Lambda^*_{X'}\mleft(0\mright)$ with respect to $q$ is always positive, establishing the independent case. Then, we extend the same argument to the case of correlated classifiers, before decoupling $q_b$ and $q_w$ for the fully correlated case from Section \ref{sec:corr_label}, setting $q_b = q_w + \delta$ for $\delta\geq 0$ based on assumption \ref{ass:qs}. Noting that by the previous proof, the theorem is correct for $\delta=0$, we again establish consistently positive derivatives of $-m \Lambda^*_X\mleft(0\mright) = -\Lambda^*_{X'}\mleft(0\mright)$, this time with respect to $\delta$. Finally, after establishing $-\Lambda^*_X\mleft(0\mright)>-\Lambda^*_{X'}\mleft(0\mright),$ and thus the first half of the theorem statement, we use assumption \ref{ass:jensen} to reduce the case of heterogeneous label accuracies $q(x)$ to the homogeneous case via stochastic dominance.
\ifthenelse{\boolean{workshop}}{Further details can be found in Appendix \ref{app:Proof}. }{
We continue with additional details for the independent case: Setting \[d \coloneqq \epsilon \mleft(1-p-\epsilon\mright)p +\mleft(\mleft(1-p-\epsilon\mright)p\mright)^2,\] 
 \[f^*\mleft(q\mright) \coloneqq M_m\mleft(q\mright) \mleft(1-M_m\mleft(q\mright)\mright) \epsilon^2 + d, \] \[g^*\mleft(q\mright) \coloneqq q \mleft(1-q\mright) \epsilon^2 + d,\] \[c \coloneqq 1 - \epsilon - 2 p \mleft(1-p-\epsilon\mright),\] it is possible to rewrite
 \begin{align*} 
& -\Lambda_{X}^*\mleft(0\mright)  + \Lambda_{X'}^*\mleft(0\mright) \ifthenelse{\boolean{icml}}{\nonumber \\&}{} = \log \mleft(2\sqrt{f^*\mleft(q\mright)}+c\mright) - m \log\mleft(2\sqrt{g^*\mleft(q\mright)}+c\mright). \end{align*}
such that 
 \begin{align*}& \frac{d}{dq}\mleft(-\Lambda_{X}^*\mleft(0\mright)  + \Lambda_{X'}^*\mleft(0\mright)\mright)  \ifthenelse{\boolean{icml}}{\nonumber \\&}{} = 
 \frac{f^{*'}\mleft(q\mright)}{\mleft(2f^*\mleft(q\mright)+c\sqrt{f^*\mleft(q\mright)}\mright)}  - m   \frac{g^{*'}\mleft(q\mright)}{\mleft(2g^*\mleft(q\mright)+c \sqrt{g^*\mleft(q\mright)} \mright)} .\end{align*} This is positive whenever 
\begin{align} \label{eq:main_main} 
f^{*'}\mleft(q\mright) \geq m  g^{*'}\mleft(q\mright)  \frac{\mleft(2f^*\mleft(q\mright)+c\sqrt{f^*\mleft(q\mright)}\mright)}{\mleft(2g^*\mleft(q\mright)+c\sqrt{g^*\mleft(q\mright)}\mright)}. 
\end{align}
To establish inequality \eqref{eq:main_main}, we calculate \[g^{*'}\mleft(q\mright) = \epsilon^2 \mleft(1-2q\mright)\] 
and 
\begin{align*}
f^{*'}\mleft(q\mright) = \epsilon^2 \mleft(1-2M_m\mleft(q\mright)\mright) m  \binom{m-1}{\frac{m-1}{2}} q^\frac{m-1}{2} \mleft(1-q\mright)^\frac{m-1}{2},
\end{align*} 
where the last equation uses the well known equality of \[M_{2n+1}\mleft(q\mright)= \mleft(2n+1\mright)  \binom{2n}{n} \int_0^q x^{n} \mleft(1-x\mright)^n dx\] \citep{boland1989modelling}. Using various algebraic manipulations to get rid of additive constants, this allows us to reduce \eqref{eq:main_main} to 
\begin{align} \label{eq:main_final}
 & \frac{ 2M_m\mleft(q\mright)-1}{2q-1} \binom{m-1}{\frac{m-1}{2}} q^\frac{m+1}{2} \mleft(1-q\mright)^\frac{m+1}{2}  \ifthenelse{\boolean{icml}}{\nonumber \\&}{}
 \leq  M_m\mleft(q\mright)\mleft(1-M_m\mleft(q\mright)\mright).
 \end{align} 
 Using an additive recursion for $M_m(q)$, we then show that both sides of the inequality approach zero for $q\rightarrow 1$. We conclude with a series of further algebraic manipulations to establish that the right hand side has a smaller derivative than the left hand side, thus growing faster as $q$ is decreased starting from $q=1$.  

The proof of the general case again centers around equation \eqref{eq:main_main}, now interpreted as a function of $\delta$. However, $f^{*}(\delta)$ and $g^{*}(\delta)$ become substantially more complicated, as they now involve both $M_m(q_w)$ and $M_m(q_w+\delta)$ terms that need to be treated separately. The proof again makes heavy use of the additive recursion for $M_m(q)$, as well as algebraic manipulations that simplify inequalities of fractions by allowing us to ignore certain terms, eventually reducing equation \eqref{eq:main_main} to equation \eqref{eq:main_final} again. For the sake of brevity, we defer further details to Appendix \ref{app:Proof}. }

\section{Conclusion}\label{sec:conclusion}
Our results suggest that while collecting multiple labels per instance can be useful for better understanding disagreement about a classification task, collecting a single label per instance is optimal for comparing binary classifiers' accuracy in terms of the annotators' majority label. Thus, while we agree with \citet{aroyo2015truth} that "one [label] is enough" is a myth when it comes to a fine-grained understanding of annotator labels, we find that one label is all you need for simple benchmarking, where a model's performance is \textit{for better or worse}, reduced to its test accuracy.

In order to better understand ambiguities in their task definition and how annotators' identity influences their labels \cite{denton2021whose}, we still encourage practitioners to initially collect multiple annotations for a small sample of instances when designing a new benchmark based on crowdsourced labels. This understanding can then be used to adjust the task instructions and annotator pool such that the expected annotator label for each instance reflects the intended task as well as possible, and in particular such that $q>0.5$ across the dataset and labels are truly noisy rather than biased \citep{reidsma2008squibs}. Achieving that might require a data-dependent annotator pool, preferentially assigning annotators to instances for which they possess relevant expertise.

Once the task description and annotator pool are fixed and it comes to evaluation at scale, we generally recommend practitioners to build their test set using a large number of instances with a single label each, according to their budget. The only exceptions are if a) estimating the precise risk $\mathcal{R}\mleft(c\mright)$ of a classifier $c$ is more important than ranking classifiers, b) the cost of unlabeled data is not negligible, or c) there is good reason to believe that one of our assumptions is violated, i.e. label errors are more common when the better classifier $c_b$ is correct or there is substantially more heterogeneity in $q\mleft(x\mright)$ when the worse classifier $c_w$ is correct. In the latter case, using single labels can still often be preferable, and we provide a calculator for the exact probabilities at \href{https://labelnoise.is.tuebingen.mpg.de}{https://labelnoise.is.tuebingen.mpg.de}. 

While we do not study the effects of aggregation for datasets that have already been constructed using multiple labels per instance, we would like to reiterate \citet{denton2021whose}'s recommendation to "Consider what valuable information might be lost through such aggregation". If such a dataset is, privacy permitting, released with all annotators' labels, users have the choice whether and how to aggregate labels \cite{prabhakaran2021releasing}. If only majority labels are released, it is impossible for others to obtain information about annotator disagreement, or even simply use a different aggregation method more suited to their needs. 

\ifthenelse{\boolean{workshop}}{}{Our work opens up multiple theoretical problems. First, while we consistently observe the single label approach outperform $m>1$ in our experiments, our main theorem is asymptotic. We conjecture, that this is always true:
\begin{conjecture}
For $G$ defined as in Section \ref{sec:corr_label} with $m>1$, \begin{align*}\ifthenelse{\boolean{icml}}{\nonumber &}{} \Pr\mleft(\sum^n_i G_i\mleft(M_m\mleft(q\mright),p\mright)>0\mright)\ifthenelse{\boolean{icml}}{\nonumber \\&}{}<\Pr\mleft(\sum^{mn}_i G_i\mleft(q,p\mright)>0\mright)\end{align*} for all $n>0$ as long as assumption \ref{ass:qs} holds. Under assumption \ref{ass:jensen}, this implies that the single label strategy outperforms the $m-$label strategy.
\end{conjecture}
Proving this conjecture likely requires different methods than employed in the current paper, as Cramér's theorem is not particularly tight for small $n$. 

Second, as our proofs involve a series of non-tight inequalities, assumptions 
\ref{ass:qs} and \ref{ass:jensen} could likely be further relaxed at the cost of additional complexity. 

Third, while binary classification is at the heart of many contemporary human-labeled tasks, most notably reward modelling for Reinforcement Learning from Human Feedback (RLHF) \citep{ouyang2022training}, multiclass classification remains an important task. Extending our results to that setting is a challenging open problem. Solving this likely requires precise modelling of class-conditional error probabilities and results might depend on details of the aggregation procedure: For example, when there are more than three labelers, plurality and absolute majority can diverge, and it is conceivable that plurality voting could extract sufficient additional signal to make collecting multiple labels competitive in some scenarios. Similarly, smarter \textit{adaptive} labeling strategies, like first collecting two labels and only collecting a third in case of a tie, could make collecting multiple labels more competitive in the binary case, but these strategies are harder to implement and analyse. }

\ifthenelse{\boolean{workshop}}{\bibliographystyle{iclr2024_conference}}{\bibliographystyle{icml2024}}

\section*{Acknowledgements}
We would like to thank mathoverflow user Kostya\_I for pointing out the connection of our problem to large deviation theory. We would also like to thank Rediet Abebe, Amin Charusaie, André Cruz, Mila Gorecki, Vivian Nastl, Olawale Salaudeen, Ana-Andreea Stoica, Sven Wang, and Jiduan Wu for helpful discussions and feedback on draft versions of this work. Florian Dorner is grateful for financial support from the Max Planck ETH Center for Learning Systems (CLS).

\ifthenelse{\boolean{icml}}{\section*{Impact Statement} This paper presents work whose goal is to advance the field of Machine Learning. There are many potential societal consequences of our work, none which we feel must be specifically highlighted here. That said, we would like to reiterate, that using single labels is only optimal for classifier comparison, once it has been established that each annotator label is more likely to be correct than not. If that is not the case, benchmark results can easily become misleading and in the worst case anti-correlated with actual task performance. As discussed in Section \ref{sec:conclusion}, collecting multiple labels for single data points during benchmark conceptualization and analyzing different annotators' disagreements can play an important role for better understanding the task definition, improving annotator instructions, and ensuring a sufficiently diverse and representative annotator pool. 
}{}

\bibliography{bib}

@inproceedings{sheng2008get,
  title={Get another label? improving data quality and data mining using multiple, noisy labelers},
  author={Sheng, Victor S and Provost, Foster and Ipeirotis, Panagiotis G},
  booktitle={Proceedings of the 14th ACM SIGKDD international conference on Knowledge discovery and data mining},
  pages={614--622},
  year={2008}
}

@book{klenke2013probability,
  title={Probability theory: a comprehensive course},
  author={Klenke, Achim},
  year={2013},
  publisher={Springer Science \& Business Media}
}

@article{boland1989modelling,
  title={Modelling dependence in simple and indirect majority systems},
  author={Boland, Philip J and Proschan, Frank and Tong, Yung Liang},
  journal={Journal of Applied Probability},
  volume={26},
  number={1},
  pages={81--88},
  year={1989},
  publisher={Cambridge University Press}
}

@inproceedings{gordon2022jury,
  title={Jury learning: Integrating dissenting voices into machine learning models},
  author={Gordon, Mitchell L and Lam, Michelle S and Park, Joon Sung and Patel, Kayur and Hancock, Jeff and Hashimoto, Tatsunori and Bernstein, Michael S},
  booktitle={Proceedings of the 2022 CHI Conference on Human Factors in Computing Systems},
  pages={1--19},
  year={2022}
}

@article{chen2021clean,
  title={Clean or annotate: How to spend a limited data collection budget},
  author={Chen, Derek and Yu, Zhou and Bowman, Samuel R},
  journal={arXiv preprint arXiv:2110.08355},
  year={2021}
}

@article{stanica2001good,
  title={Good lower and upper bounds on binomial coefficients},
  author={Stanica, Pantelimon},
  journal={Journal of Inequalities in Pure and Applied Mathematics},
  volume={2},
  number={3},
  pages={30},
  year={2001}
}

@article{ouyang2022training,
  title={Training language models to follow instructions with human feedback},
  author={Ouyang, Long and Wu, Jeffrey and Jiang, Xu and Almeida, Diogo and Wainwright, Carroll and Mishkin, Pamela and Zhang, Chong and Agarwal, Sandhini and Slama, Katarina and Ray, Alex and others},
  journal={Advances in Neural Information Processing Systems},
  volume={35},
  pages={27730--27744},
  year={2022}
}

@article{russakovsky2015imagenet,
  title={Imagenet large scale visual recognition challenge},
  author={Russakovsky, Olga and Deng, Jia and Su, Hao and Krause, Jonathan and Satheesh, Sanjeev and Ma, Sean and Huang, Zhiheng and Karpathy, Andrej and Khosla, Aditya and Bernstein, Michael and others},
  journal={International journal of computer vision},
  volume={115},
  pages={211--252},
  year={2015},
  publisher={Springer}
}

@article{kopf2023openassistant,
  title={OpenAssistant Conversations--Democratizing Large Language Model Alignment},
  author={K{\"o}pf, Andreas and Kilcher, Yannic and von R{\"u}tte, Dimitri and Anagnostidis, Sotiris and Tam, Zhi-Rui and Stevens, Keith and Barhoum, Abdullah and Duc, Nguyen Minh and Stanley, Oliver and Nagyfi, Rich{\'a}rd and others},
  journal={arXiv preprint arXiv:2304.07327},
  year={2023}
}

@article{dorner2022human,
  title={Human-Guided Fair Classification for Natural Language Processing},
  author={Dorner, Florian E and Peychev, Momchil and Konstantinov, Nikola and Goel, Naman and Ash, Elliott and Vechev, Martin},
  journal={arXiv preprint arXiv:2212.10154},
  year={2022}
}

@article{touvron2023llama,
  title={Llama 2: Open foundation and fine-tuned chat models},
  author={Touvron, Hugo and Martin, Louis and Stone, Kevin and Albert, Peter and Almahairi, Amjad and Babaei, Yasmine and Bashlykov, Nikolay and Batra, Soumya and Bhargava, Prajjwal and Bhosale, Shruti and others},
  journal={arXiv preprint arXiv:2307.09288},
  year={2023}
}

@inproceedings{fleisig2023fair,
  title={Fair-Prism: Evaluating fairness-related harms in text generation},
  author={Fleisig, Eve and Amstutz, Aubrie and Atalla, Chad and Blodgett, Su Lin and Daum{\'e} III, Hal and Olteanu, Alexandra and Sheng, Emily and Vann, Dan and Wallach, Hanna},
  booktitle={Proceedings of the 61st Annual Meeting of the Association for Computational Linguistics. Association for Computational Linguistics},
  year={2023}
}

@article{davani2022dealing,
  title={Dealing with disagreements: Looking beyond the majority vote in subjective annotations},
  author={Davani, Aida Mostafazadeh and D{\'\i}az, Mark and Prabhakaran, Vinodkumar},
  journal={Transactions of the Association for Computational Linguistics},
  volume={10},
  pages={92--110},
  year={2022},
  publisher={MIT Press One Rogers Street, Cambridge, MA 02142-1209, USA journals-info~…}
}

@inproceedings{sandri2023don,
  title={Why don’t you do it right? analysing annotators’ disagreement in subjective tasks},
  author={Sandri, Marta and Leonardelli, Elisa and Tonelli, Sara and Je{\v{z}}ek, Elisabetta},
  booktitle={Proceedings of the 17th Conference of the European Chapter of the Association for Computational Linguistics},
  pages={2420--2433},
  year={2023}
}

@inproceedings{cheplygina2018crowd,
  title={Crowd disagreement about medical images is informative},
  author={Cheplygina, Veronika and Pluim, Josien PW},
  booktitle={Intravascular Imaging and Computer Assisted Stenting and Large-Scale Annotation of Biomedical Data and Expert Label Synthesis: 7th Joint International Workshop, CVII-STENT 2018 and Third International Workshop, LABELS 2018, Held in Conjunction with MICCAI 2018, Granada, Spain, September 16, 2018, Proceedings 3},
  pages={105--111},
  year={2018},
  organization={Springer}
}

@article{aroyo2013crowd,
  title={Crowd truth: Harnessing disagreement in crowdsourcing a relation extraction gold standard},
  author={Aroyo, Lora and Welty, Chris},
  journal={WebSci2013. ACM},
  volume={2013},
  number={2013},
  year={2013}
}

@misc{krizhevsky2009learning,
  title={Learning multiple layers of features from tiny images},
  author={Krizhevsky, Alex and Hinton, Geoffrey and others},
  year={2009},
  publisher={Toronto, ON, Canada}
}

@inproceedings{tanno2019learning,
  title={Learning from noisy labels by regularized estimation of annotator confusion},
  author={Tanno, Ryutaro and Saeedi, Ardavan and Sankaranarayanan, Swami and Alexander, Daniel C and Silberman, Nathan},
  booktitle={Proceedings of the IEEE/CVF conference on computer vision and pattern recognition},
  pages={11244--11253},
  year={2019}
}

@inproceedings{recht2019imagenet,
  title={Do imagenet classifiers generalize to imagenet?},
  author={Recht, Benjamin and Roelofs, Rebecca and Schmidt, Ludwig and Shankar, Vaishaal},
  booktitle={International conference on machine learning},
  pages={5389--5400},
  year={2019},
  organization={PMLR}
}

@inproceedings{blum2015ladder,
  title={The ladder: A reliable leaderboard for machine learning competitions},
  author={Blum, Avrim and Hardt, Moritz},
  booktitle={International Conference on Machine Learning},
  pages={1006--1014},
  year={2015},
  organization={PMLR}
}

@article{crammer2005learning,
  title={Learning from data of variable quality},
  author={Crammer, Koby and Kearns, Michael and Wortman, Jennifer},
  journal={Advances in Neural Information Processing Systems},
  volume={18},
  year={2005}
}

@book{gray2019ghost,
  title={Ghost work: How to stop Silicon Valley from building a new global underclass},
  author={Gray, Mary L and Suri, Siddharth},
  year={2019},
  publisher={Eamon Dolan Books}
}

@article{bowman2015large,
  title={A large annotated corpus for learning natural language inference},
  author={Bowman, Samuel R and Angeli, Gabor and Potts, Christopher and Manning, Christopher D},
  journal={arXiv preprint arXiv:1508.05326},
  year={2015}
}

@inproceedings{socher2013recursive,
  title={Recursive deep models for semantic compositionality over a sentiment treebank},
  author={Socher, Richard and Perelygin, Alex and Wu, Jean and Chuang, Jason and Manning, Christopher D and Ng, Andrew Y and Potts, Christopher},
  booktitle={Proceedings of the 2013 conference on empirical methods in natural language processing},
  pages={1631--1642},
  year={2013}
}

@article{williams2017broad,
  title={A broad-coverage challenge corpus for sentence understanding through inference},
  author={Williams, Adina and Nangia, Nikita and Bowman, Samuel R},
  journal={arXiv preprint arXiv:1704.05426},
  year={2017}
}

@inproceedings{marelli2014semeval,
  title={Semeval-2014 task 1: Evaluation of compositional distributional semantic models on full sentences through semantic relatedness and textual entailment},
  author={Marelli, Marco and Bentivogli, Luisa and Baroni, Marco and Bernardi, Raffaella and Menini, Stefano and Zamparelli, Roberto},
  booktitle={Proceedings of the 8th international workshop on semantic evaluation (SemEval 2014)},
  pages={1--8},
  year={2014}
}

@inproceedings{dolan2005automatically,
  title={Automatically constructing a corpus of sentential paraphrases},
  author={Dolan, Bill and Brockett, Chris},
  booktitle={Third International Workshop on Paraphrasing (IWP2005)},
  year={2005}

}

@article{nguyen2022vindr,
  title={VinDr-CXR: An open dataset of chest X-rays with radiologist’s annotations},
  author={Nguyen, Ha Q and Lam, Khanh and Le, Linh T and Pham, Hieu H and Tran, Dat Q and Nguyen, Dung B and Le, Dung D and Pham, Chi M and Tong, Hang TT and Dinh, Diep H and others},
  journal={Scientific Data},
  volume={9},
  number={1},
  pages={429},
  year={2022},
  publisher={Nature Publishing Group UK London}
}

@misc{Jigsaw,
 author = {Jigsaw},
  title = {{Jigsaw} Unintended Bias in Toxicity Classification},
  howpublished = {\url{https://www.kaggle.com/competitions/jigsaw-unintended-bias-in-toxicity-classification/data}},
  year={2019},
  note = {Accessed: 2024-01-24}}

@inproceedings{wei2023aggregate,
  title={To aggregate or not? learning with separate noisy labels},
  author={Wei, Jiaheng and Zhu, Zhaowei and Luo, Tianyi and Amid, Ehsan and Kumar, Abhishek and Liu, Yang},
  booktitle={Proceedings of the 29th ACM SIGKDD Conference on Knowledge Discovery and Data Mining},
  pages={2523--2535},
  year={2023}
}

@inproceedings{ramponi2022dh,
  title={DH-FBK at SemEval-2022 task 4: leveraging annotators’ disagreement and multiple data views for patronizing language detection},
  author={Ramponi, Alan and Leonardelli, Elisa},
  booktitle={Proceedings of the 16th International Workshop on Semantic Evaluation (SemEval-2022)},
  pages={324--334},
  year={2022}
}

@inproceedings{lin2014re,
  title={To re (label), or not to re (label)},
  author={Lin, Christopher and Weld, Daniel and others},
  booktitle={Proceedings of the AAAI Conference on Human Computation and Crowdsourcing},
  volume={2},
  pages={151--158},
  year={2014}
}

@article{mania2020classifier,
  title={Why do classifier accuracies show linear trends under distribution shift?},
  author={Mania, Horia and Sra, Suvrit},
  journal={arXiv preprint arXiv:2012.15483},
  year={2020}
}

@book{hardtrecht2022patterns,
  author = {Moritz Hardt and Benjamin Recht},
  title = {Patterns, predictions, and actions: Foundations of machine learning},
  year = {2022},
  publisher = {Princeton University Press}
}

@article{aroyo2015truth,
  title={Truth is a lie: Crowd truth and the seven myths of human annotation},
  author={Aroyo, Lora and Welty, Chris},
  journal={AI Magazine},
  volume={36},
  number={1},
  pages={15--24},
  year={2015}
}

@article{denton2021whose,
  title={Whose ground truth? accounting for individual and collective identities underlying dataset annotation},
  author={Denton, Emily and D{\'\i}az, Mark and Kivlichan, Ian and Prabhakaran, Vinodkumar and Rosen, Rachel},
  journal={arXiv preprint arXiv:2112.04554},
  year={2021}
}

@article{prabhakaran2021releasing,
  title={On releasing annotator-level labels and information in datasets},
  author={Prabhakaran, Vinodkumar and Davani, Aida Mostafazadeh and Diaz, Mark},
  journal={arXiv preprint arXiv:2110.05699},
  year={2021}
}

@article{cheng2022many,
  title={How many labelers do you have? A closer look at gold-standard labels},
  author={Cheng, Chen and Asi, Hilal and Duchi, John},
  journal={arXiv preprint arXiv:2206.12041},
  year={2022}
}

@article{cheng2025some,
  title={Some Robustness Properties of Label Cleaning},
  author={Cheng, Chen and Duchi, John},
  journal={arXiv preprint arXiv:2509.11379},
  year={2025}
}

@article{reidsma2008squibs,
  title={Squibs: Reliability measurement without limits},
  author={Reidsma, Dennis and Carletta, Jean},
  journal={Computational linguistics},
  volume={34},
  number={3},
  pages={319--326},
  year={2008}
}
\appendix
\onecolumn
\section{Numerical Evidence}\label{sec:experiments}
We conducted a large scale parameter sweep for \[n\in [1,2,3,4,5,6,7,8,9,10,11,100,101,1000,1001],\] \[m\in [3,11]\] and \[q_b,q_w,p_w,p_b^0,p_b^1 \in S^{5},\] where $S$ is a set of $50$ evenly spaced points $s\in [0.5,0.99]$ with a resolution of $0.01$. For all of the almost five billion grid points that fulfilled $ \mleft(1-p_w\mright) p_b^0 + p_w p_b^1  > p_w$, we both explicitly calculated 
\[\Pr\mleft(\sum_{i=1}^{mn} G_i\mleft(q,p,\epsilon\mright)>0\mright)\] and \[\Pr\mleft(\sum_{i=1}^n G_i\mleft(M_{m}\mleft(q\mright),p,\epsilon\mright)>0\mright)\] (using iterated convolutions of the base variable $G$, sped up via exponentiation by squaring) and additionally approximated the probabilities based on sampling each of the sums $100$ times. 
Under the assumptions from section \ifthenelse{\boolean{workshop}}{\ref{sec:base}}{\ref{sec:corr_label}}, the exact calculations consistently yielded \[\Pr\mleft(\sum_{i=1}^{mn} G_i\mleft(q,p,\epsilon\mright)>0\mright)\geq \Pr\mleft(\sum_{i=1}^n G_i\mleft(M_{m}\mleft(q\mright),p,\epsilon\mright)>0\mright),\] with the only exceptions happening when both probabilities are extremely close to $1$ (maximal distance of the order $1e-12$). These exceptions do not provide meaningful evidence against our conjecture, as they are most likely caused by numerical instability (notably, they often coincide with calculated probabilities that exceed one). In particular, there were no parameters for which both the exact probabilities and the sampled probabilities were better for the $m-$label case, even though this happened for the sampled probabilities alone in $1.6\%$ of the cases (as to be expected from the relatively small sample size of $100$). As an additional sanity check, the sampled probabilities generally approximated the exact probabilities well, with the average distance over all parameters being on the order of $1e-7$, and the average MSE of the order $0.01$ for both the single and the $m-$label case. 
 
Notably, the single label approach still performed better in two thirds of the parameter configurations with $q_w>q_b$, with this number slowly decreasing for larger values of $q_w$. This suggests that our (already not particularly restrictive) assumptions could be relaxed substantially further. 

\section{Parameterizations of the Gap Indicator}\label{app:params}
\ifthenelse{\boolean{workshop}}{\begin{restatable}{prop}{propindep}
Assuming mutually independent classifier and labeler errors (i.e. $p_w=p$, $p_b^0=p_b^1 = p + \epsilon$, $q_w=q_b =q$), $G$ can be written as follows:
\[
    G\mleft(q,p,\epsilon\mright) = 
    \begin{cases}
    1 &\text{ w.p. } q\epsilon + \mleft(1-p-\epsilon\mright)p  \\
    -1 &\text{ w.p. } \mleft(1-q\mright)\epsilon  + \mleft(1-p-\epsilon\mright)p  \\
    0 &\text{ else }  p\mleft(p+\epsilon\mright) + \mleft(1-p-\epsilon\mright)\mleft(1-p\mright) 
    \end{cases},
\]
for label accuracy $q$, classifier accuracy $p$ and margin $\epsilon$. 
\end{restatable}}{\propindep*}
\begin{proof}

The better classifier $c_b$ wins for a given $x$ (i.e. $G=1$) if $c_b\mleft(x\mright)$ and the label $y_{\mathit{Test}}\mleft(x\mright)$ are correct, while $c_w\mleft(x\mright)$ is not, or if both $c_b\mleft(x\mright)$ and the label $y_{\mathit{Test}}\mleft(x\mright)$ are incorrect, while $c_w\mleft(x\mright)$ is correct. The former happens with probability $\mleft(\mleft(p+\epsilon\mright)\mleft(1-p\mright)q\mright)$, and the latter with probability $\mleft(\mleft(p\mright)\mleft(1-p-\epsilon\mright)\mleft(1-q\mright)\mright)$. Summing up yields 
\begin{align*}
   \Pr\mleft(G=1\mright) = & \mleft(\mleft(p+\epsilon\mright)\mleft(1-p\mright)q\mright) + \mleft(p\mright)\mleft(1-p-\epsilon\mright)\mleft(1-q\mright) \\ & = 
   qp - qp^2 +q\epsilon - qp \epsilon + \mleft(1-q\mright) \mleft(p-p^2-p\epsilon\mright) \\ & = 
   qp - qp^2 +q\epsilon - qp \epsilon + p-p^2-p\epsilon - qp +qp^2 + qp\epsilon \\ & = 
   q \epsilon + p - p^2 - p\epsilon =  q \epsilon + p \mleft(1-p-\epsilon\mright).
\end{align*} For the worse classifier $c_w$ to win ($G=-1$), we get the opposite cases conditional on the label, with respective probabilities of $\mleft(\mleft(p+\epsilon\mright)\mleft(1-p\mright)\mleft(1-q\mright)\mright)$ and $(\mleft(p\mright)\mleft(1-p-\epsilon\mright)q$. These sum up as follows:
\begin{align*}
    \Pr\mleft(G=-1\mright) &= \mleft(\mleft(p+\epsilon\mright)\mleft(1-p\mright)\mleft(1-q\mright)\mright) + \mleft(p\mright)\mleft(1-p-\epsilon\mright)q \\ & = 
   \mleft(p + \epsilon - p^2 - p\epsilon\mright) \mleft(1-q\mright) + qp - qp^2 -qp\epsilon \\ & =
   p + \epsilon - p^2 - p\epsilon -qp -q\epsilon +qp^2 +qp\epsilon + qp - qp^2 -qp\epsilon \\ & = p +\epsilon -p^2 - p\epsilon - q\epsilon \\& = 
   \mleft(1-p-\epsilon\mright)p+\mleft(1-q\mright)\epsilon. 
\end{align*}
Adding up both probabilities yields 
\begin{align*}
\Pr\mleft(G\neq 0\mright) &= 2 p \mleft(1-p-\epsilon\mright) + \epsilon \\ & = 
2p - 2p^2 - 2p\epsilon + \epsilon \\&=
1 - p \mleft(p+\epsilon\mright) + 2p - p^2 - p\epsilon +\epsilon -1 
\\&= 1 - p\mleft(p+\epsilon\mright) - \mleft(1-p-\epsilon\mright)\mleft(1-p\mright), 
\end{align*}
which makes sense as the gap indicator $G\mleft(p,q,\epsilon\mright)$ is zero whenever both classifiers produce the same answer, independent of the label. 
\end{proof}
\ifthenelse{\boolean{workshop}}{\begin{restatable}{prop}{propdeplabel}
Assuming correlated classifiers and labels with the above parameterization, we have:
\[
    G\mleft(q,p\mright) = 
    \begin{cases} 
    1 &\text{w.p. } q_b \mleft(1-p_w\mright) p_b^0 \ifthenelse{\boolean{icml}}{\nonumber \\&}{}+  \mleft(1-q_w\mright) p_w \mleft(1-p_b^1\mright)\\
    -1 &\text{w.p. }  \mleft(1-q_b\mright) \mleft(1-p_w\mright) p_b^0 \ifthenelse{\boolean{icml}}{\nonumber \\&}{}+  q_w p_w \mleft(1-p_b^1\mright) \\
    0 &\text{else }  
    \end{cases}.
\]
\end{restatable}}
{\propdeplabel*}
\begin{proof}
    The better classifier $c_b$ "wins" on a given datapoint, whenever it and the label are correct, while the worse classifier is not, or if the label and the better classifier are incorrect, while the worse classifier is correct. The former happens with probability $q_b \mleft(1-p_w\mright) p_b^0 $ and the latter with probability $\mleft(1-q_w\mright) p_w \mleft(1-p_b^1\mright)$. The case of the worse classifier winning is symmetric, with $q_i$ and $1-q_i$ reversed. This yields 
\[
    G\mleft(q_b,q_w,p_w,p_b^0,p_b^1\mright) = 
    \begin{cases}
    1 &\text{ w.p. } q_b \mleft(1-p_w\mright) p_b^0 +  \mleft(1-q_w\mright) p_w \mleft(1-p_b^1\mright)\\
    -1 &\text{ w.p. } \mleft(1-q_b\mright) \mleft(1-p_w\mright) p_b^0 +  q_w p_w \mleft(1-p_b^1\mright) \\
    0 &\text{ else }  
    \end{cases}.
\]
\end{proof}

\ifthenelse{\boolean{workshop}}{}{
\section{Details on Hoeffding Bounds}\label{app:hoeffding}
We first establish, that it is sufficient to focus on the case of $m$ uneven, as going from $m$ uneven to $m+1$ even reduces the number of data points we can label $n$, while \textit{reducing} $M_m\mleft(q\mright)$ due to additional ties, rather than increasing it: 
\begin{lemma}
For even $k>1$, we have that
\begin{align*}&M_k\mleft(q\mright) < M_{k-1}\mleft(q\mright). 
\end{align*}
\end{lemma}
\begin{proof}
For even $k$, a majority can only be obtained, if there is already a majority for the first $k-1$ votes. In that case, the majority is always retained, unless the margin was exactly one, and the new vote goes against the majority. For our case, this means that \begin{align*}&M_k\mleft(q\mright) = M_{k-1}\mleft(q\mright) - \binom{k-1}{\frac{k}{2}}q^{\frac{k}{2}} \mleft(1-q\mright)^{\frac{k}{2}} < M_{k-1}\mleft(q\mright). 
\end{align*}
\end{proof}
We proceed by proving lemma \ref{lemma:hoeff}:

\ifthenelse{\boolean{workshop}}
{\begin{restatable}{lemma}{lemmahoeffding}\label{lemma:hoeff}
For independent copies $X_i$ of any random variable $X$ with $\E[X]>0$ and values in $[-1,1]$, we can bound \[\Pr\mleft(\sum^n_{i=1} X_i \leq 0\mright) \leq  e^{ \frac{- n \E[X]^2 }{2}}\eqqcolon B(X,n).\] 
\end{restatable}}{\lemmahoeffding*}
\begin{proof}
    For independent copies $X_i$ of any random variable $X$ with values in $[-1,1]$, we have \begin{align*}
    \sum^n_{i=1} X_i \leq 0 & \iff
    \sum^n_{i=1} \mleft(X_i - \E[X]\mright) \leq -n \E[X]  \iff 
     \sum^n_{i=1} \mleft(- X_i  - \E[-X]\mright) \geq n \E[X].
\end{align*}
If $\E[X]>0$, we can then apply Hoeffding's inequality to $-X$ to obtain \[\Pr\mleft(\sum^n_{i=1} X_i \leq 0\mright) = \Pr\mleft(\sum^n_{i=1} \mleft(- X_i  - \E[-X]\mright) \geq n \E[X]\mright) \leq e^{ \frac{-2 n^2 \E[X]^2 }{4n}} =   e^{ \frac{- n \E[X]^2 }{2}}.\] 
\end{proof}
With this, we focus on
\begin{restatable}{prop}{theoremhoeffdingbase} \label{thm:hoeffding_base}
For any uneven $m>1$, equation \eqref{hoeffding} is true, i.e. \[
\sqrt{m}   >  \frac{2M_m\mleft(q\mright)-1 }{2q-1}. \] Correspondingly, for any $m>1$ \[ B\mleft(G\mleft(q,p,\epsilon\mright),nm\mright) < B\mleft(G\mleft(M_{m}\mleft(q\mright),p,\epsilon\mright),n\mright),\] where $B$ is the Hoeffding lower bound on the success probability. 
\end{restatable}
\begin{proof}
To prove proposition \ref{thm:hoeffding_base}, we need the following lemma:
\begin{lemma} \label{lemma:sigma}
    Setting $\sigma\mleft(m,q\mright)=\sum^{m-2}_{k  \text{ uneven}} \binom{k}{\ceil{\frac{k}{2}}} q^{\ceil{\frac{k}{2}}} \mleft(1-q\mright)^{\ceil{\frac{k}{2}}} $ for uneven $m$, we have that \[M_m\mleft(q\mright) = q + \mleft(2q-1\mright) \sigma\mleft(m,q\mright).\]
\end{lemma}
\begin{proof}
    Let $b_n\mleft(q,k\mright)$ be the probability of $k$ successes in a binomial with $n$ trials and with probability of success $q$ for a single trial. Then: 
\begin{align*}
& M_m\mleft(q\mright) = M_{m-2}\mleft(q\mright) + q^2  b_{m-2}\mleft(q,\floor{\frac{m-2}{2}}\mright) - \mleft(1-q\mright)^2 b_{m-2}\mleft(q,\ceil{\frac{m-2}{2}}\mright) \\ & =
M_{m-2}\mleft(q\mright) + q^2 \frac{1-q}{q} b_{m-2}\mleft(q,\ceil{\frac{m-2}{2}}\mright) - \mleft(1-q\mright)^2 b_{m-2}\mleft(q,\ceil{\frac{m-2}{2}}\mright)
\\ & = M_{m-2}\mleft(q\mright) + \mleft(q - q^2 - 1 + 2q  - q^2\mright) b_{m-2}\mleft(q,\ceil{\frac{m-2}{2}}\mright)
\\ & = M_{m-2}\mleft(q\mright) + \mleft(3q - 2q^2 -1\mright) b_{m-2}\mleft(q,\ceil{\frac{m-2}{2}}\mright)
\\ & = M_{m-2}\mleft(q\mright) + \mleft(1-q\mright) \mleft(2q-1\mright) b_{m-2}\mleft(q,\ceil{\frac{m-2}{2}}\mright)
\\ & = M_{m-2}\mleft(q\mright) + \mleft(1-q\mright) \mleft(2q-1\mright) \binom{m-2}{\ceil{\frac{m-2}{2}}} q^{\ceil{\frac{m-2}{2}}} \mleft(1-q\mright)^{\ceil{\frac{m-2}{2}}-1}
\\ & = M_{m-2}\mleft(q\mright) + \mleft(2q-1\mright)  \binom{m-2}{\ceil{\frac{m-2}{2}}} q^{\ceil{\frac{m-2}{2}}} \mleft(1-q\mright)^{\ceil{\frac{m-2}{2}}}.\end{align*}
The first equation captures the fact that a majority of $m$ trials consists of all events that have a majority for the first $m-2$ trials (first term), except for those with a margin of one that simultaneously have two misses in the last two trials (third term), in addition to all events that miss a majority in the first $m-2$ trials by a margin of one, but have two successes in the last two trials (second term). The statement of the Lemma then follows by unrolling the additive recursion.
\end{proof}
With Lemma \ref{lemma:sigma}, \eqref{hoeffding} can be rewritten as \begin{equation}\label{hoeffding_sigma}
    \sqrt{m}   >  \frac{2M_m\mleft(q\mright)-1 }{2q-1} = \frac{2q + 2\mleft(2q-1\mright) \sigma\mleft(m,q\mright) -1}{2q-1} = 1 + 2 \sigma\mleft(m,q\mright).
\end{equation}
We can control the right term using another Lemma:
\begin{lemma} \label{lemma:sqrtbound}
    \[1+ 2\sigma\mleft(m,q\mright) \leq  1 + \frac{1}{\sqrt{\pi}} (2 \sqrt{\frac{m-1}{2}} -1)  \] 
\end{lemma}
\begin{proof}
We use an upper bound version of Stirling's approximation based on Theorem 2.6 in \citep{stanica2001good}: \[\binom{m-1}{\frac{m-1}{2}} < \frac{4^\frac{m-1}{2}}{\sqrt{\pi \frac{m-1}{2}}},\] the fact that $q\mleft(1-q\mright)$ is maximized at $q=0.5$ and the monotonicity of $\frac{1}{\sqrt{k}}$
to estimate 
\begin{alignat}{3}
  2\sigma\mleft(m,q\mright) \nonumber
  &=   2 \sum^{m-2}_{k  \text{ uneven}} \binom{k}{\ceil{\frac{k}{2}}} q^{\ceil{\frac{k}{2}}} \mleft(1-q\mright)^{\ceil{\frac{k}{2}}} && =
  2 \sum^{m-2}_{k  \text{ uneven}}\binom{k}{\frac{k+1}{2}} q^{\frac{k+1}{2}} \mleft(1-q\mright)^{\frac{k+1}{2}} 
  \nonumber
 \\& = 
   2 \sum^{m-2}_{k  \text{ uneven}} \frac{\frac{k+1}{2}}{k+1} \binom{k+1}{\frac{k+1}{2}} q^{\frac{k+1}{2}} \mleft(1-q\mright)^{\frac{k+1}{2}} 
   &&  =
     \sum^{m-2}_{k  \text{ uneven}} \binom{k+1}{\frac{k+1}{2}} q^{\frac{k+1}{2}} \mleft(1-q\mright)^{\frac{k+1}{2}} 
  \nonumber
\\& \leq
    \sum^{m-2}_{k  \text{ uneven}}  \frac{4^\frac{k+1}{2}}{\sqrt{\pi \frac{k+1}{2}}} q^{\frac{k+1}{2}} \mleft(1-q\mright)^{\frac{k+1}{2}}   &&
   \leq
    \sum^{m-2}_{k  \text{ uneven}}  \frac{1}{\sqrt{\pi \frac{k+1}{2}}} 
   \nonumber
  \\&  =  \frac{1}{\sqrt{\pi}} \sum^{m-1}_{k>0  \text{ even}}  \frac{1}{\sqrt{ \frac{k}{2}}}
  && =   \frac{1}{\sqrt{\pi}} \sum^{\frac{m-1}{2}}_{k=1}  \frac{1}{\sqrt{k}}
 \nonumber
\\& =   \frac{1}{\sqrt{\pi}} \mleft(1+ \sum^{\frac{m-1}{2}}_{k=2}  \frac{1}{\sqrt{k}}\mright)
 &&\leq   \frac{1}{\sqrt{\pi}} \mleft(1+ \int^{\frac{m-1}{2}}_{k=1}  \frac{1}{\sqrt{k}}\mright)
 \nonumber
  \\& =   \frac{1}{\sqrt{\pi}} \mleft(1 + 2 \sqrt{\frac{m-1}{2}} -2\mright)
  && =   \frac{1}{\sqrt{\pi}} \mleft( 2 \sqrt{\frac{m-1}{2}} -1\mright)
  &\nonumber.
\end{alignat}
\end{proof}
With this, \eqref{hoeffding} reduces to \[1 - \frac{1}{\sqrt{\pi}} + \sqrt{\frac{2}{\pi} \mleft(m-1\mright)}   = 1 + \frac{1}{\sqrt{\pi}} \mleft(2 \sqrt{\frac{m-1}{2}} -1\mright) < \sqrt{m}.\]  At $m=3$, this becomes \[ 1.56 \approx 1 + \frac{1}{\sqrt{\pi}} < \sqrt{3} \approx 1.73.\] On the other hand the derivative of the gap with respect to $m$, \[\frac{d}{dm} \mleft(\sqrt{m} - 1 + \frac{1}{\sqrt{\pi}} - \sqrt{\frac{2}{\pi} \mleft(m-1\mright)}  \mright) = \frac{1}{2 \sqrt{m}} - \frac{1}{\sqrt{2 \pi} \sqrt{m-1}}\] is positive whenever \[ \frac{1}{2 \sqrt{m}} >  \frac{1}{\sqrt{2 \pi} \sqrt{m-1}}\] or \[ 1.25 \approx \frac{\sqrt{2 \pi} }{2 } > \sqrt{\frac{m}{m-1}} = \sqrt{1 + \frac{1}{m-1}}.
\]
For $m\geq 3$, the right side is clearly at most $\sqrt{1 + \frac{1}{3-1}} \approx 1.22$, such that the derivative is positive for all $m>3$ and \eqref{hoeffding} holds for $m\geq 3$. 
\end{proof}
Next, we focus on the general case with correlated classifiers and labels:
\begin{restatable}{prop}{theoremhoeffdingcorrlabel}  \label{thm:hoeffding_corr_label}
When classifiers and labels are correlated, as long as $q_b\geq q_w$ and $ \mleft(1-p_w\mright) p_b^0 + p_w p_b^1  > p_w$, \[ B\mleft(G\mleft(q,p\mright),nm\mright) < B\mleft(G\mleft(M_{m}\mleft(q\mright),p\mright),n\mright)\] holds for any $m>1$,  where $B$ is the Hoeffding lower bound on the success probability. 
\end{restatable}

\begin{proof} In this setting, 
Equation \eqref{hoeffding_abstract} becomes \begin{equation}\label{hoeffding_corr}
\sqrt{m}   >  \frac{\mleft(2M_m\mleft(q_b\mright)-1\mright) \mleft(1-p_w\mright) p_b^0 - \mleft(2M_m\mleft(q_w\mright)-1\mright)  p_w \mleft(1-p_b^1\mright) }{\mleft(2q_b-1\mright) \mleft(1-p_w\mright) p_b^0 - \mleft(2q_w-1\mright)  p_w \mleft(1-p_b^1\mright)}. 
\end{equation} 
To prove this, we need the following Lemma:
\begin{restatable}{lemma}{lemmamagicreverse}\label{lemma:magic_reverse}
    Let $A,B,C,D,c_1,c_2$ be positive constants such that $Ac_1 - Bc_2 > 0$ and $ Cc_1-Dc_2  > 0$.
    Then $\frac{ Ac_1-Bc_2 }{ Cc_1 - Dc_2}\leq \frac{A}{C}$ is true if and only if $CB\geq DA$.
\end{restatable}
\begin{proof}
    \begin{align*}
        \frac{ Ac_1- Bc_2}{ Cc_1 - Dc_2}\leq \frac{A}{C}  & \iff
         Ac_1 - Bc_2\leq \frac{A\mleft( Cc_1 -Dc_2 \mright)}{C}  \\& \iff 
          C\mleft(Ac_1 - Bc_2\mright)\leq A\mleft(Cc_1  - Dc_2\mright)  \\& \iff 
           C A c_1 - C  B c_2 \leq  C A c_1 - D A c_2 \\& \iff 
           - C  B c_2  \leq  - D A c_2  \\& \iff 
            C  B\geq   D A 
    \end{align*}
\end{proof} 
With this, we set
\[A=2M_m\mleft(q_b\mright)-1,\] \[B=2M_m\mleft(q_w\mright)-1,\] \[C=2q_b-1,\] \[D=2q_w-1,\] and \[
c_1 = \mleft(1-p_w\mright) p_b^0,\] \[c_2 = p_w \mleft(1-p_b^1\mright).\] 
Then, $CB\geq DA$ is equivalent to \[\mleft(2q_b-1\mright)\mleft(2M_m\mleft(q_w\mright)-1\mright)\geq \mleft(2q_w-1\mright)\mleft(2M_m\mleft(q_b\mright)-1\mright),\]  i.e.  \[\frac{2M_m\mleft(q_w\mright)-1}{2q_w-1}\geq  \frac{2M_m\mleft(q_b\mright)-1}{2q_b-1},\] which is equivalent to 
\[1+ 2\sigma\mleft(m,q_w\mright)\geq  1+ 2\sigma\mleft(m,q_b\mright),\] and holds for $q_b\geq q_w$ as $\sigma\mleft(m,x\mright)$ is clearly monotonically decreasing in $x$. 
Lemma \ref{lemma:magic_reverse} combined with Equation \eqref{hoeffding_sigma} thus allows us to upper bound 
\begin{align*}
      \frac{\mleft(2M_m\mleft(q_b\mright)-1\mright) \mleft(1-p_w\mright) p_b^0 - \mleft(2M_m\mleft(q_w\mright)-1\mright)  p_w \mleft(1-p_b^1\mright) }{\mleft(2q_b-1\mright) \mleft(1-p_w\mright) p_b^0 - \mleft(2q_w-1\mright)  p_w \mleft(1-p_b^1\mright)}  \leq \frac{2M_m\mleft(q_b\mright)-1}{2q_b -1}  \leq \sqrt{m},
\end{align*} proving the proposition.  
\end{proof}}
\section{Proving Theorem \ref{thm:asym_optim}} \label{app:Proof}
\main* 
The proof of theorem \ref{thm:asym_optim} is based on Cramér's Theorem: 
\ifthenelse{\boolean{workshop}}{
\begin{restatable}{theorem_cramer}{cramer} (Adapted from \citep{klenke2013probability}) Let $X_i$ be iid real random variables for $i \in \mathbb{N}$ such that \[\Lambda\mleft(t\mright)\coloneqq\log \E[e^{tX_{1}}]<\infty\] for all $t \in \mathbb{R}$. 
Define the Legendre transform \[\Lambda^*\mleft(x\mright)\coloneqq\sup_t \mleft(tx-\Lambda\mleft(t\mright)\mright).\] Then for all $z\in \mathbb{R}$ such that $z>\E[X_1]$, we have \[\lim_{n\to\infty}\frac{1}{n} \log \Pr\mleft(S_n=\sum_{i=1}^n X_i\geq zn\mright) = - \Lambda^*\mleft(z\mright),\] where the limit is an upper bound for all $n$.  
\end{restatable}}{\cramer*}
This means that $\Pr\mleft(S_n=\sum_{i=1}^n X_i\geq zn\mright)$ is eventually roughly of the order $e^{- n\Lambda^*\mleft(z\mright)}$. Furthermore, a glance at the proof of Cramér's Theorem, reveals that this exponential is actually an upper bound for the error probability independent of $n$ in our case of $z=0$. We want to eventually apply the theorem to $X = -G\mleft(M_m\mleft(q\mright),p,\epsilon\mright)$ and $X' = -\sum_{i=1}^m G_i\mleft(q,p,\epsilon\mright)$ respectively. Because these random variables have negative expectation, the theorem can be applied to $z=0>\E[X]$, yielding limits for 
\begin{align*}\frac{1}{n} \log \Pr\mleft(S_n\geq 0\mright) \coloneqq \frac{1}{n} \log \mleft( \Pr\mleft(\sum^n_i G_i\mleft(M_m\mleft(q\mright),p,\epsilon\mright)\leq 0\mright)\mright) = \frac{1}{n} \log \mleft( 1-\Pr\mleft(\sum^n_i G_i\mleft(M_m\mleft(q\mright),p,\epsilon\mright)>0\mright)\mright)  \end{align*} and \[
\frac{1}{n} \log \Pr\mleft(S'_n\geq 0\mright) \coloneqq \frac{1}{n} \log \mleft( 1-\Pr\mleft(\sum^{mn}_i G_i\mleft(q,p,\epsilon\mright)>0\mright)\mright).
\]
If we can prove that 
\begin{equation}\label{Cramer_main}-\Lambda^*_X\mleft(0\mright)>-\Lambda^*_{X'}\mleft(0\mright),\end{equation}it follows that there is an $N\in \mathbb{N}$ such that for $n>N$ we have \begin{align*}\frac{1}{n} \log \mleft( 1-\Pr\mleft(\sum^n_i G_i\mleft(M_m\mleft(q\mright),p,\epsilon\mright)>0\mright)\mright)  >\frac{1}{n} \log \mleft( 1-\Pr\mleft(\sum^{mn}_i G_i\mleft(q,p,\epsilon\mright)>0\mright)\mright) \end{align*} and thus by monotonicity \[\Pr\mleft(\sum^n_i G_i\mleft(M_m\mleft(q\mright),p,\epsilon\mright)>0\mright)<\Pr\mleft(\sum^{mn}_i G_i\mleft(q,p,\epsilon\mright)>0\mright).\] We first consider a general ternary $X$ with negative expectation: \[
    X = 
    \begin{cases}
    1 &\text{ w.p. } x  \\
    -1 &\text{ w.p. } y  \\
    0 &\text{ w.p. }  z
    \end{cases}
\] for $y>x$ .

\ifthenelse{\boolean{workshop}}{
\begin{restatable}{lemma}{ternary}\label{lemma:explicit_lambda}
For $X$ ternary with $\Pr\mleft(X=1\mright) = x$, $\Pr\mleft(X=-1\mright) = y$, and $\Pr\mleft(X=0\mright) = z$, \[-\Lambda_X^*\mleft(0\mright)=\log{\mleft(2\sqrt{xy}+z\mright)}.\] For sums of independent copies $X_i$ of ternary variables $S_n=\sum_{i=1}^n X_i,$ \[-\Lambda_{S_n}^*\mleft(0\mright)=n\log{\mleft(2\sqrt{xy}+z\mright)}.\]
\end{restatable}}{\ternary*}
\begin{proof}
We have that \[-\Lambda_X^*\mleft(0\mright)= - \mleft(\sup_t 0\cdot t -\Lambda_X\mleft(t\mright)\mright) =  \inf_t \Lambda_X\mleft(t\mright) \]  Here,  \[ \Lambda_X\mleft(t\mright) = \log{\E[e^{tX}]}=  \log\mleft(x e^t + y e^{-t} +z\mright).\]  Differentiating yields \[\frac{d}{dt}\Lambda_X\mleft(t\mright) = \frac{x e^t - y e^{-t} }{x e^t + y e^{-t} +z}.\]  
The numerator is positive for large positive $t$ and negative for large negative $t$, with a unique zero at $xe^t = ye^{-t}$, i.e. $\frac{y}{x}=e^{2t}$ or $t = 0.5 \log{\frac{y}{x}}$, such that $\Lambda\mleft(t\mright)$ is minimized at this t. 
This means that \begin{align*}\inf_t \Lambda_X\mleft(t\mright) &= \log\mleft(x e^{0.5 \log{\frac{y}{x}}} + y e^{-0.5 \log{\frac{y}{x}}} +z\mright) \\&=\log\mleft(x \sqrt{e^{\log{\frac{y}{x}}}} + y \frac{1}{\sqrt{e^{ \log{\frac{y}{x}}}} }+z\mright) \\&= \log\mleft(x \sqrt{\frac{y}{x}} + y \sqrt{\frac{x}{y}} +z\mright) \\&= \log \mleft(2 \sqrt{xy} + z\mright).\end{align*}
Now for $S_n$, we get \[ \Lambda_{S_n}\mleft(t\mright) = \log{\E[e^{t Sn}]} = \log{\prod_{i=1}^n\E[e^{t X_i}]} =  n \log\mleft(x e^t + y e^{-t} +z\mright).\] The optimization is not affected by multiplying by $n$, so we get \[\inf_t \Lambda_{S_n}\mleft(t\mright) =  n\log \mleft(2 \sqrt{xy} + z\mright)\]
\end{proof} 
\subsection{Independent Classifiers}\label{sec:ld}
We first focus on the independent case with \[
    G\mleft(q,p,\epsilon\mright) = 
    \begin{cases}
    1 &\text{ w.p. } q\epsilon + \mleft(1-p-\epsilon\mright)p  \\
    -1 &\text{ w.p. } \mleft(1-q\mright)\epsilon  + \mleft(1-p-\epsilon\mright)p  \\
    0 &\text{ else }  p\mleft(p+\epsilon\mright) + \mleft(1-p-\epsilon\mright)\mleft(1-p\mright) 
    \end{cases},
\] where $q_b=q_w=q$, $p_w=p$ and $p_b^0=p_b^1 = p+\epsilon$ as defined in section \ref{sec:base} and apply Lemma \ref{lemma:explicit_lambda} to \[X=-G\mleft(M_m\mleft(q\mright),p,\epsilon\mright)\] and \[X' = -\sum_{i=1}^m G_i\mleft(q,p,\epsilon\mright)\] 
to obtain \begin{align*}-\Lambda_X^*\mleft(0\mright) =   \log\biggl(2 \sqrt{M_m\mleft(q\mright) \mleft(1-M_m\mleft(q\mright)\mright) \epsilon^2 + \epsilon \mleft(1-p-\epsilon\mright)p +\mleft(\mleft(1-p-\epsilon\mright)p\mright)^2}  + 1 - \epsilon - 2 p \mleft(1-p-\epsilon\mright)\biggr).\end{align*} and 
\begin{align*}-\Lambda_{X'}^*\mleft(0\mright) =   m \log\biggl(2 \sqrt{q \mleft(1-q\mright) \epsilon^2 + \epsilon \mleft(1-p-\epsilon\mright)p +\mleft(\mleft(1-p-\epsilon\mright)p\mright)^2} + 1 - \epsilon - 2 p \mleft(1-p-\epsilon\mright)\biggr).\end{align*} 
To get some intuition, we fix $p=0.5$, such that 
\[ -\Lambda_X^*\mleft(0\mright) = \log\mleft( {2 \sqrt{\mleft(M_m\mleft(q\mright) \mleft(1-M_m\mleft(q\mright)\mright)- \frac{1}{4}\mright)\epsilon^2 + \frac{1}{16}}   } + \frac{1}{2}\mright),\] which for the aggregated case yields an asymptotic error rate of 
\[ e^ {-\Lambda_X^*\mleft(0\mright)n} =   \mleft(2 \sqrt{\mleft(M_m\mleft(q\mright) \mleft(1-M_m\mleft(q\mright)\mright)- \frac{1}{4}\mright)\epsilon^2 + \frac{1}{16}}     + \frac{1}{2}\mright)^n 
\]
The error rate has a second order Taylor expansion around $\epsilon=0$ of \[e^{-n\Lambda_X^*\mleft(0\mright)} \approx 1+ \mleft(4\mleft(M_m\mleft(q\mright) \mleft(1-M_m\mleft(q\mright)\mright)-1\mright)\mright) n\epsilon^2.\] For $q=M_m\mleft(q\mright)=1,$ we thus get \[e^{-n\Lambda_X^*\mleft(0\mright)} \approx 1 - n\epsilon^2,\] which is consistent with the statistical intuition that $n\gg\frac{1}{\epsilon^2}$ samples are needed to detect a coin with a bias of order $\epsilon$.  Meanwhile as $q$ goes to $0.5$, $M_m\mleft(q\mright) \mleft(1-M_m\mleft(q\mright)\mright)$ approaches $4$ and the amount of required samples explodes. 

Back to general $p\geq 0.5$, we note that by the AM-GM inequality, $2 \sqrt{xy} + z \leq x+y+z=1$ for any $x,y,z$ that describe a ternary random variable as above with equality only if $x=y$, which cannot happen for $G$ under our assumptions because of its positive expectation. This means that the logarithms in the $\Lambda^*$ are always strictly negative. In particular, at $q=0.5$ and $q=1$, $M_m\mleft(q\mright)=q$ such that the terms in the logarithm are equal and we get $-\Lambda_X^*\mleft(0\mright) \geq -m\Lambda_X^*\mleft(0\mright) = -\Lambda_{X'}^*\mleft(0\mright)$. In general, we have 
\begin{align} \label{Cramer_Explicit}
  &-\Lambda_{X}^*\mleft(0\mright)  + \Lambda_{X'}^*\mleft(0\mright)\\&= \log \biggl(2 \sqrt{M_m\mleft(q\mright) \mleft(1-M_m\mleft(q\mright)\mright) \epsilon^2 + \epsilon \mleft(1-p-\epsilon\mright)p +\mleft(\mleft(1-p-\epsilon\mright)p\mright)^2} \nonumber  + 1 - \epsilon - 2 p \mleft(1-p-\epsilon\mright)\biggr) \\ & - m \log  \biggl(2 \sqrt{q \mleft(1-q\mright) \epsilon^2 + \epsilon \mleft(1-p-\epsilon\mright)p +\mleft(\mleft(1-p-\epsilon\mright)p\mright)^2} \nonumber   +1 - \epsilon - 2 p \mleft(1-p-\epsilon\mright)\biggr).
\end{align} 
Because \eqref{Cramer_main} holds for $q=0.5$ independent of $\epsilon$ and $p$ as both terms are the same except for the factor $m$ in that case, it is sufficient to show that \eqref{Cramer_Explicit} always has a positive derivative in $q$. To show this, we set 
 \[f^*\mleft(q\mright) = M_m\mleft(q\mright) \mleft(1-M_m\mleft(q\mright)\mright) \epsilon^2 + \epsilon \mleft(1-p-\epsilon\mright)p +\mleft(\mleft(1-p-\epsilon\mright)p\mright)^2, \] \[g^*\mleft(q\mright) = q \mleft(1-q\mright) \epsilon^2 + \epsilon \mleft(1-p-\epsilon\mright)p +\mleft(\mleft(1-p-\epsilon\mright)p\mright)^2 \] and \[c = 1 - \epsilon - 2 p \mleft(1-p-\epsilon\mright),\] such that 
 \begin{equation} \label{Cramer_Reduced}
-\Lambda_{X}^*\mleft(0\mright)  + \Lambda_{X'}^*\mleft(0\mright) = \log \mleft(2\sqrt{f^*\mleft(q\mright)}+c\mright) - m \log\mleft(2\sqrt{g^*\mleft(q\mright)}+c\mright). \end{equation}
 Differentiating yields 
 \begin{align*}\frac{d}{dq}\mleft(-\Lambda_{X}^*\mleft(0\mright)  + \Lambda_{X'}^*\mleft(0\mright)\mright) & = \frac{d}{dq} \log \mleft(2\sqrt{f^*\mleft(q\mright)}+c\mright) - m  \frac{d}{dq}  \log\mleft(2\sqrt{g^*\mleft(q\mright)}+c\mright)
 \\& = \frac{\frac{d}{dq} 2\sqrt{f^*\mleft(q\mright)}}{\mleft(2\sqrt{f^*\mleft(q\mright)}+c\mright)} - m \frac{\frac{d}{dq} 2\sqrt{g^*\mleft(q\mright)}}{\mleft(2\sqrt{g^*\mleft(q\mright)}+c\mright)} \\& = 
 \frac{\frac{f^{*'}\mleft(q\mright)}{\sqrt{f^*\mleft(q\mright)}}}{\mleft(2\sqrt{f^*\mleft(q\mright)}+c\mright)}  - m  \frac{\frac{g^{*'}\mleft(q\mright)}{\sqrt{g^*\mleft(q\mright)}}}{\mleft(2\sqrt{g^*\mleft(q\mright)}+c\mright)} 
 \\& = 
 \frac{f^{*'}\mleft(q\mright)}{\sqrt{f^*\mleft(q\mright)}\mleft(2\sqrt{f^*\mleft(q\mright)}+c\mright)} \\& - m   \frac{g^{*'}\mleft(q\mright)}{\sqrt{g^*\mleft(q\mright)}\mleft(2\sqrt{g^*\mleft(q\mright)}+c\mright)} 
 .\end{align*}
Correspondingly using \eqref{Cramer_Explicit}, \eqref{Cramer_main} reduces to 
\begin{equation}\label{Cramer_Final}
f^{*'}\mleft(q\mright) \geq m  g^{*'}\mleft(q\mright)  \frac{\sqrt{f^*\mleft(q\mright)}\mleft(2\sqrt{f^*\mleft(q\mright)}+c\mright)}{\sqrt{g^*\mleft(q\mright)}\mleft(2\sqrt{g^*\mleft(q\mright)}+c\mright)} 
\end{equation}
We can calculate \[g^{*'}\mleft(q\mright) = \frac{d}{dq} q\mleft(1-q\mright) \epsilon^2 = \epsilon^2 \mleft(1-2q\mright)\] 
and 
\begin{align*}
f^{*'}\mleft(q\mright)  &=  \epsilon^2  \frac{d}{dq} M_m\mleft(q\mright)\mleft(1-M_m\mleft(q\mright)\mright) \\& = 
\epsilon^2 \mleft(1-2M_m\mleft(q\mright)\mright) \frac{d}{dq} M_m\mleft(q\mright)
\\& = \epsilon^2 \mleft(1-2M_m\mleft(q\mright)\mright) m  \binom{m-1}{\frac{m-1}{2}} q^\frac{m-1}{2} \mleft(1-q\mright)^\frac{m-1}{2},
\end{align*} 
where the last equation uses the equality of \[M_{2n+1}\mleft(q\mright)= \mleft(2n+1\mright)  \binom{2n}{n} \int_0^q x^{n} \mleft(1-x\mright)^n dx\] \citep{boland1989modelling}. Correspondingly, \eqref{Cramer_Final} holds if and only if \begin{align*} \mleft(1-2M_m\mleft(q\mright)\mright) m  \binom{m-1}{\frac{m-1}{2}} q^\frac{m-1}{2} \mleft(1-q\mright)^\frac{m-1}{2}
 \geq m \mleft(1-2q\mright) \frac{\sqrt{f^*\mleft(q\mright)}\mleft(2\sqrt{f^*\mleft(q\mright)}+c\mright)}{\sqrt{g^*\mleft(q\mright)}\mleft(2\sqrt{g^*\mleft(q\mright)}+c\mright)} .
\end{align*}
For $0.5<q<1$, this is equivalent to 
 \begin{equation}\label{prestirling}
 \frac{2M_m\mleft(q\mright)-1}{2q-1} \binom{m-1}{\frac{m-1}{2}} q^\frac{m-1}{2} \mleft(1-q\mright)^\frac{m-1}{2}
\leq \frac{\sqrt{f^*\mleft(q\mright)}\mleft(2\sqrt{f^*\mleft(q\mright)}+c\mright)}{\sqrt{g^*\mleft(q\mright)}\mleft(2\sqrt{g^*\mleft(q\mright)}+c\mright)}.
 \end{equation}

\begin{lemma}\label{lemma:fraction}
    Let $0<x<y$ and $c>0$. Then, $\frac{2x+c\sqrt{x}}{2y+c\sqrt{y}}\geq \frac{x}{y}$
\end{lemma}
\begin{proof}
\begin{align*}
      \frac{2x+c\sqrt{x}}{2y+c\sqrt{y}}\geq \frac{x}{y}  & \iff
     \mleft(2x  + \sqrt{x}c\mright) y \geq  \mleft(2y  + \sqrt{y}c\mright) x \\ & \iff
     2xy + c \sqrt{x} y \geq 2xy + c\sqrt{y} x 
    \\ & \iff  \sqrt{x} y \geq \sqrt{y} x 
    \\ & \iff \frac{y}{\sqrt{y}} \geq \frac{x}{\sqrt{x}} 
    \\ & \iff \sqrt{y} \geq \sqrt{x} 
     \\ & \iff y \geq x
\end{align*}
\end{proof}
As $f^*\mleft(q\mright)$ and $g^*\mleft(q\mright)$ can be written as $M_m\mleft(q\mright)\mleft(1-M_m\mleft(q\mright)\mright)\epsilon^2 + d$ and  $q\mleft(1-q\mright)\epsilon^2 + d$ respectively for $d=\epsilon \mleft(1-p-\epsilon\mright)p +\mleft(\mleft(1-p-\epsilon\mright)p\mright)^2$, and because $k\mleft(x\mright)=x\mleft(1-x\mright)$ is monotonically falling in $x$ while $M_m\mleft(x\mright)$ grows in $m$, $g^*\mleft(x\mright) \geq f^*\mleft(x\mright)$, and Lemma \ref{lemma:fraction} implies that it is sufficient to show  
\begin{equation}\label{prestirling2}
 \frac{2M_m\mleft(q\mright)-1}{2q-1} \binom{m-1}{\frac{m-1}{2}} q^\frac{m-1}{2} \mleft(1-q\mright)^\frac{m-1}{2}
\leq \frac{f^*\mleft(q\mright)}{g^*\mleft(q\mright)}.
 \end{equation}
\begin{lemma}\label{lemma:fraction2}
    Let $0<x<y$ and $d>0$. Then, $\frac{x+d}{y+d}\geq \frac{x}{y}$
\end{lemma}
\begin{proof}
    \begin{align*}
     \frac{x+d}{y+d}\geq \frac{x}{y} & \iff
        y\mleft(x+d\mright) \geq x\mleft(y+d\mright) \\ & \iff
        xy + yd \geq yx + xd \\ & \iff
        y \geq x 
    \end{align*}
\end{proof}
Lemma \ref{lemma:fraction2} implies that \eqref{prestirling2} can be reduced to  
\begin{equation}\label{prestirling3}
 \frac{ 2M_m\mleft(q\mright)-1}{2q-1} \binom{m-1}{\frac{m-1}{2}} q^\frac{m-1}{2} \mleft(1-q\mright)^\frac{m-1}{2}
\leq \frac{\epsilon^2 M_m\mleft(q\mright)\mleft(1-M_m\mleft(q\mright)\mright)}{ \epsilon^2 q\mleft(1-q\mright)}.
 \end{equation}
\ifthenelse{\boolean{workshop}}{
Next, we need the following lemma:
\begin{lemma} \label{lemma:sigma}
    Setting $\sigma\mleft(m,q\mright)=\sum^{m-2}_{k  \text{ uneven}} \binom{k}{\ceil{\frac{k}{2}}} q^{\ceil{\frac{k}{2}}} \mleft(1-q\mright)^{\ceil{\frac{k}{2}}} $ for uneven $m$, we have that \[M_m\mleft(q\mright) = q + \mleft(2q-1\mright) \sigma\mleft(m,q\mright).\]
\end{lemma}
\begin{proof}
    Let $b_n\mleft(q,k\mright)$ be the probability of $k$ successes in a binomial with $n$ trials and with probability of success $q$ for a single trial. Then: 
\begin{align*}
& M_m\mleft(q\mright) = M_{m-2}\mleft(q\mright) + q^2  b_{m-2}\mleft(q,\floor{\frac{m-2}{2}}\mright) - \mleft(1-q\mright)^2 b_{m-2}\mleft(q,\ceil{\frac{m-2}{2}}\mright) \\ & =
M_{m-2}\mleft(q\mright) + q^2 \frac{1-q}{q} b_{m-2}\mleft(q,\ceil{\frac{m-2}{2}}\mright) - \mleft(1-q\mright)^2 b_{m-2}\mleft(q,\ceil{\frac{m-2}{2}}\mright)
\\ & = M_{m-2}\mleft(q\mright) + \mleft(q - q^2 - 1 + 2q  - q^2\mright) b_{m-2}\mleft(q,\ceil{\frac{m-2}{2}}\mright)
\\ & = M_{m-2}\mleft(q\mright) + \mleft(3q - 2q^2 -1\mright) b_{m-2}\mleft(q,\ceil{\frac{m-2}{2}}\mright)
\\ & = M_{m-2}\mleft(q\mright) + \mleft(1-q\mright) \mleft(2q-1\mright) b_{m-2}\mleft(q,\ceil{\frac{m-2}{2}}\mright)
\\ & = M_{m-2}\mleft(q\mright) + \mleft(1-q\mright) \mleft(2q-1\mright) \binom{m-2}{\ceil{\frac{m-2}{2}}} q^{\ceil{\frac{m-2}{2}}} \mleft(1-q\mright)^{\ceil{\frac{m-2}{2}}-1}
\\ & = M_{m-2}\mleft(q\mright) + \mleft(2q-1\mright)  \binom{m-2}{\ceil{\frac{m-2}{2}}} q^{\ceil{\frac{m-2}{2}}} \mleft(1-q\mright)^{\ceil{\frac{m-2}{2}}}.\end{align*}
The first equation captures the fact that a majority of $m$ trials consists of all events that have a majority for the first $m-2$ trials (first term), except for those with a margin of one that simultaneously have two misses in the last two trials (third term), in addition to all events that miss a majority in the first $m-2$ trials by a margin of one, but have two successes in the last two trials (second term). The statement of the Lemma then follows by unrolling the additive recursion.
\end{proof}
 }{}
Lemma \ref{lemma:sigma} allows to rewrite \eqref{prestirling3} as 
\begin{align*}
& \mleft(1 + 2 \sigma\mleft(m,q\mright)\mright) \binom{m-1}{\frac{m-1}{2}} q^\frac{m-1}{2} \mleft(1-q\mright)^\frac{m-1}{2} 
\\&= \frac{2q + 2\mleft(2q-1\mright) \sigma\mleft(m,q\mright)-1}{2q-1}  \binom{m-1}{\frac{m-1}{2}} q^\frac{m-1}{2} \mleft(1-q\mright)^\frac{m-1}{2}
\\& \leq \frac{M_m\mleft(q\mright)\mleft(1-M_m\mleft(q\mright)\mright)}{q\mleft(1-q\mright)}
 \end{align*}
 or equivalently 
 \begin{equation} \label{prederivative}
 \mleft(1 + 2 \sigma\mleft(m,q\mright)\mright) \binom{m-1}{\frac{m-1}{2}} q^\frac{m+1}{2} \mleft(1-q\mright)^\frac{m+1}{2}  \leq  M_m\mleft(q\mright)\mleft(1-M_m\mleft(q\mright)\mright). \end{equation} We note that both sides approach zero from above as $q\rightarrow 1$, such that \eqref{prederivative} holds for $q=1$. It is thus sufficient to show, that the right side grows faster than the left side when decreasing $q$, i.e. \begin{equation} \label{derivative_base}
      \frac{d}{dq}\mleft(\mleft(1 + 2 \sigma\mleft(m,q\mright)\mright) \binom{m-1}{\frac{m-1}{2}} q^\frac{m+1}{2} \mleft(1-q\mright)^\frac{m+1}{2}\mright) \geq \frac{d}{dq}\mleft(M_m\mleft(q\mright)\mleft(1-M_m\mleft(q\mright)\mright)\mright).
 \end{equation}
We have 
\begin{align*}
     \frac{d}{dq} \mleft(M_m\mleft(q\mright)\mleft(1-M_m\mleft(q\mright)\mright)\mright) & =  \mleft(1-2M_m\mleft(q\mright)\mright) m \binom{m-1}{\frac{m-1}{2}} q^\frac{m-1}{2} \mleft(1-q\mright)^\frac{m-1}{2} 
\end{align*}
Meanwhile, 
\begin{align*}
     & \frac{d}{dq}  \mleft(1 + 2 \sigma\mleft(m,q\mright)\mright) \binom{m-1}{\frac{m-1}{2}} q^\frac{m+1}{2} \mleft(1-q\mright)^\frac{m+1}{2} \\& = 
     \mleft(1+2\sigma\mleft(m,q\mright)\mright) \frac{m+1}{2} \mleft(1-2q\mright) \binom{m-1}{\frac{m-1}{2}}  q^\frac{m-1}{2} \mleft(1-q\mright)^\frac{m-1}{2} + \binom{m-1}{\frac{m-1}{2}}  q^\frac{m-1}{2} \mleft(1-q\mright)^\frac{m-1}{2} q\mleft(1-q\mright) 2 \frac{d}{dq} \sigma\mleft(m,q\mright)
     \\& =  \binom{m-1}{\frac{m-1}{2}}  q^\frac{m-1}{2} \mleft(1-q\mright)^\frac{m-1}{2}  \mleft( 
     \mleft(1+2\sigma\mleft(m,q\mright)\mright) \frac{m+1}{2} \mleft(1-2q\mright) + q\mleft(1-q\mright) 2 \frac{d}{dq} \sigma\mleft(m,q\mright)\mright).
\end{align*}
Because $\binom{m-1}{\frac{m-1}{2}}  q^\frac{m-1}{2} \mleft(1-q\mright)^\frac{m-1}{2}>0$ for $q<1$, \eqref{derivative_base} or \[
\frac{d}{dq}  \mleft(1 + 2 \sigma\mleft(m,q\mright)\mright) \binom{m-1}{\frac{m-1}{2}} q^\frac{m+1}{2} \mleft(1-q\mright)^\frac{m+1}{2}  \geq \frac{d}{dq} \mleft(M_m\mleft(q\mright)\mleft(1-M_m\mleft(q\mright)\mright)\mright) \] holds whenever \begin{equation} \label{derivative}
     \mleft(1+2\sigma\mleft(m,q\mright)\mright) \frac{m+1}{2} \mleft(1-2q\mright) + q\mleft(1-q\mright) 2\frac{d}{dq} \sigma\mleft(m,q\mright) \geq  m\mleft(1-2M_m\mleft(q\mright)\mright).
\end{equation}
Dividing by the (negative) $1-2M_m\mleft(q\mright)$ term yields
\[ 
     \mleft(1+2\sigma\mleft(m,q\mright)\mright) \frac{m+1}{2} \frac{\mleft(1-2q\mright) }{1-2M_m\mleft(q\mright)}+ \frac{q\mleft(1-q\mright)}{1-2M_m\mleft(q\mright)} 2\frac{d}{dq} \sigma\mleft(m,q\mright) \leq m .
\]
which is equivalent to 
\[ 
      \frac{m+1}{2} + \frac{q\mleft(1-q\mright)}{1-2M_m\mleft(q\mright)} 2\frac{d}{dq} \sigma\mleft(m,q\mright)\leq m 
\] as $\frac{\mleft(1-2q\mright) }{1-2M_m\mleft(q\mright)} = \frac{1}{1+2\sigma\mleft(m,q\mright)}$. Rewriting yields
\begin{align}\label{derivative_reduced}
     \frac{m-1}{2} &= m - \frac{m+1}{2} \\&  \geq - \frac{q\mleft(1-q\mright)}{2M_m\mleft(q\mright)-1} 2\frac{d}{dq} \sigma\mleft(m,q\mright) \nonumber
    \\ & =  - 2\frac{q\mleft(1-q\mright)}{2M_m\mleft(q\mright)-1}  \mleft(1-2q\mright)  \sum^{m-2}_{k  \text{ uneven}} \frac{k+1}{2}\binom{k}{\frac{k+1}{2}} q^{\frac{k-1}{2}} \mleft(1-q\mright)^{\frac{k-1}{2}} \nonumber
    \\ & =  2\frac{2q-1}{2M_m\mleft(q\mright)-1}   \sum^{m-2}_{k  \text{ uneven}} \frac{k+1}{2}\binom{k}{\frac{k+1}{2}} q^{\frac{k+1}{2}} \mleft(1-q\mright)^{\frac{k+1}{2}} \nonumber
    \\ & = 2\frac{1}{1+2\sigma\mleft(m,q\mright)} \sum^{m-2}_{k  \text{ uneven}} \frac{k+1}{2}\binom{k}{\frac{k+1}{2}} q^{\frac{k+1}{2}} \mleft(1-q\mright)^{\frac{k+1}{2}} 
\end{align}
We can upper bound \begin{align*} \sum^{m-2}_{k  \text{ uneven}} \frac{k+1}{2}\binom{k}{\frac{k+1}{2}} q^{\frac{k+1}{2}} \mleft(1-q\mright)^{\frac{k+1}{2}}  &\leq \frac{m-2+1}{2} \sum^{m-2}_{k  \text{ uneven}}\binom{k}{\frac{k+1}{2}} q^{\frac{k+1}{2}} \mleft(1-q\mright)^{\frac{k+1}{2}} \\& = \frac{m-1}{2} \sigma\mleft(m,q\mright)
\end{align*}
such that \eqref{derivative_reduced} reduces to \begin{equation}
\frac{m-1}{2}  \geq \frac{m-1}{2} \frac{ 2\sigma\mleft(m,q\mright)}{1+2\sigma\mleft(m,q\mright)},
\end{equation}
which is clearly true, as $\frac{x}{1+x} < \frac{x}{x}=1$ for all $x>0$. 

\subsection{Correlated Classifiers}
We now analyze the case of correlated classifiers discussed in \ifthenelse{\boolean{workshop}}{\ref{sec:base}}{\ref{sec:corr_label}}, at first keeping $q= q_b = q_w$ fixed to be equal. As a reminder, we now have 
\[
    G\mleft(q,p,\epsilon\mright) = 
    \begin{cases}
    1 &\text{ w.p. } q \mleft(1-p_w\mright) p_b^0 +  \mleft(1-q\mright) p_w \mleft(1-p_b^1\mright)\\
    -1 &\text{ w.p. } \mleft(1-q\mright) \mleft(1-p_w\mright) p_b^0 +  q p_w \mleft(1-p_b^1\mright) \\
    0 &\text{ else }  
    \end{cases}.
\]
with expectation \begin{align*}
& \mleft(2q-1\mright) \mleft(1-p_w\mright) p_b^0 - \mleft(2q-1\mright)  p_w \mleft(1-p_b^1\mright) \\ & = \mleft(2q-1\mright) \mleft(\mleft(1-p_w\mright) p_b^0 + p_w \mleft(p_b^1-1\mright)\mright)  > 0. 
\end{align*}
We also note that \begin{align*}
\Pr\mleft(G\mleft(q,p,\epsilon\mright)=0\mright)  &=   1-\Pr\mleft(G\mleft(q,p,\epsilon\mright)=1\mright)-\Pr\mleft(G\mleft(q,p,\epsilon\mright)=-1\mright) \\&=
1 - q \mleft(1-p_w\mright) p_b^0 -  \mleft(1-q\mright) p_w \mleft(1-p_b^1\mright) \\& - \mleft(1-q\mright) \mleft(1-p_w\mright) p_b^0 -  q p_w \mleft(1-p_b^1\mright) \\& = 
1 - \mleft(1-p_w\mright) p_b^0 - p_w \mleft(1-p_b^1\mright) \eqqcolon c_0
\end{align*} is constant in $q$.  Repeating the argument from above, we now obtain 
\begin{align*}
 & \Lambda_X'^*\mleft(0\mright) \\& = m \log \biggl( 2\sqrt{ \Pr\mleft(G\mleft(q,p,\epsilon\mright)=1\mright) \Pr\mleft(G\mleft(q,p,\epsilon\mright)=-1\mright) } + c_0 \biggr)
\\& = m \log \biggl( 2 \biggl(\mleft( q \mleft(1-p_w\mright) p_b^0 \mleft(1-q\mright) \mleft(1-p_w\mright) p_b^0 +
             q \mleft(1-p_w\mright) p_b^0 q p_w \mleft(1-p_b^1\mright)\mright) \\ &+ 
             \mleft(1-q\mright) p_w \mleft(1-p_b^1\mright) \mleft(1-q\mright) \mleft(1-p_w\mright) p_b^0 + 
              \mleft(1-q\mright) p_w \mleft(1-p_b^1\mright) q p_w \mleft(1-p_b^1\mright)\biggr)^\frac{1}{2}  + c_0  \biggr) 
     \\& = m \log \mleft(2\sqrt{q\mleft(1-q\mright)c_1 + q q c_2 + \mleft(1-q\mright)\mleft(1-q\mright) c_3 + \mleft(1-q\mright) q c_4 } +c_0\mright),
\end{align*} where the $c_i$ are constants that do not depend on $q$. 
We also note, that $p_w \mleft(1-p_b^1\mright) \mleft(1-p_w\mright) p_b^0 = c_2= c_3 $.  
 
We now consider \begin{align*}
f^*\mleft(q\mright) &=  \mleft(c_1+c_4\mright) M_m\mleft(q\mright) \mleft(1-M_m\mleft(q\mright)\mright) + c_2\mleft(M_m\mleft(q\mright)^2 + \mleft(1-M_m\mleft(q\mright)\mright)^2\mright) 
\\& =  \mleft(c_1+c_4\mright) M_m\mleft(q\mright) \mleft(1-M_m\mleft(q\mright)\mright) + c_2\mleft(M_m\mleft(q\mright)^2 + 1 - 2M_m\mleft(q\mright) + M_m\mleft(q\mright)^2\mright)
\\& =  \mleft(c_1+c_4\mright) M_m\mleft(q\mright) \mleft(1-M_m\mleft(q\mright)\mright) + c_2\mleft(2M_m\mleft(q\mright)^2  - 2M_m\mleft(q\mright)\mright) +c_2 
\\& =  \mleft(c_1+c_4\mright) M_m\mleft(q\mright) \mleft(1-M_m\mleft(q\mright)\mright) - 2c_2\mleft(M_m\mleft(q\mright) \mleft(1-M_m\mleft(q\mright)\mright)\mright) +c_2  
\\& =\mleft(c_1+c_4- 2c_2\mright) M_m\mleft(q\mright) \mleft(1-M_m\mleft(q\mright)\mright) +c_2 
\end{align*}
and  \[g^*\mleft(q\mright) =\mleft(c_1+c_4- 2c_2\mright) q\mleft(1-q\mright) +c_2 ,\]
such that \[-\Lambda_{X}^*\mleft(0\mright)  + \Lambda_{X'}^*\mleft(0\mright) = \log \mleft(2\sqrt{f^*\mleft(q\mright)}+c_0\mright) - m \log\mleft(2\sqrt{g^*\mleft(q\mright)}+c_0\mright),\] where $c_0$ does not depend on $q$. This is exactly \eqref{Cramer_Reduced} with $c_0$ replacing  $c$. A brief glance reveals that \[\frac{\mleft(1-p_w\mright)^2 \mleft(p_b^0\mright)^2 +  \mleft(1-p_b^1\mright)^2 \mleft(p_w\mright)^2}{2}  \geq \mleft(\mleft(1-p_w\mright) p_b^0 p_w \mleft(1-p_b^1\mright)\mright)\]  by the AM-GM inequality, such that \[c_1+c_4 - 2c_2> 0.\] This means that $f^*\mleft(q\mright)$ and $g^*\mleft(q\mright)$ are exactly of the form $d_1 M_m\mleft(q\mright) \mleft(1-M_m\mleft(q\mright)\mright)+d_2$ and $d_1 q \mleft(1-q\mright)+d_2$
for constants $d_1=c_1+c_4 - 2c_2>0$ and $d_2=c_2>0$. As it did not rely on the specific values for these constants beyond their positivity, the reasoning from the last section (where $d_1 = \epsilon^2$ and $d_2=\epsilon \mleft(1-p-\epsilon\mright)p +\mleft(\mleft(1-p-\epsilon\mright)p\mright)^2 $) can be repeated one to one, proving our main result for correlated classifiers,

\subsection{Correlated Classifiers and Labels}
As in \ifthenelse{\boolean{workshop}}{\ref{sec:base}}{\ref{sec:corr_label}}, we now consider 
\[
    G\mleft(q,p,\epsilon\mright) = 
    \begin{cases}
    1 &\text{ w.p. } q_b \mleft(1-p_w\mright) p_b^0 +  \mleft(1-q_w\mright) p_w \mleft(1-p_b^1\mright)\\
    -1 &\text{ w.p. } \mleft(1-q_b\mright) \mleft(1-p_w\mright) p_b^0 +  q_w p_w \mleft(1-p_b^1\mright) \\
    0 &\text{ else }  
    \end{cases}.
\]
with expectation
\begin{align*}
\mleft(2q_b-1\mright) \mleft(1-p_w\mright) p_b^0 - \mleft(2q_w-1\mright)  p_w \mleft(1-p_b^1\mright) > 0. 
\end{align*}
We note that \begin{align*}
\Pr\mleft(G\mleft(q,p,\epsilon\mright)=0\mright)  &=   1-\Pr\mleft(G\mleft(q,p,\epsilon\mright)=1\mright)-\Pr\mleft(G\mleft(q,p,\epsilon\mright)=-1\mright) \\&=
1 - q_b \mleft(1-p_w\mright) p_b^0 -  \mleft(1-q_w\mright) p_w \mleft(1-p_b^1\mright) \\& - \mleft(1-q_b\mright) \mleft(1-p_w\mright) p_b^0 -  q_w p_w \mleft(1-p_b^1\mright) \\& = 
1 - \mleft(1-p_w\mright) p_b^0 - p_w \mleft(1-p_b^1\mright)
\end{align*} still does not depend on either of the $q_i$, nor their difference. 
By assumption \ref{ass:qs}, we can reparameterise $q_b=q_w + \delta =q +\delta$ for $\delta\geq 0$ and we know by the previous calculations that \eqref{Cramer_main} holds for $\delta=0$. We now obtain 
\begin{align*}
  - \Lambda_X'^*\mleft(0\mright) 
     & =  m \log \biggl(2\biggl(\mleft(q+\delta\mright) \mleft(1-q-\delta\mright)c_1 + \mleft(q+\delta\mright) q c_2 + \mleft(1-q\mright)\mleft(1-q-\delta\mright) c_3 + \mleft(1-q\mright) q c_4 \biggr)^{\frac{1}{2}} +c_0 \biggr),
\end{align*} where the constants $c_i$ are as before and neither depend on $q$ nor $\delta$. We set 
\begin{align*}
f^*\mleft(\delta\mright) &= c_1 M_m\mleft(q+\delta\mright) \mleft(1-M_m\mleft(q+\delta\mright)\mright) + c_2\biggl(M_m\mleft(q\mright)M_m\mleft(q+\delta\mright)   +\mleft(1-M_m\mleft(q\mright)\mright)\mleft(1-M_m\mleft(q+\delta\mright)\mright) \biggr) \\ & + c_4 \mleft(1-M_m\mleft(q\mright)\mright)M_m\mleft(q\mright)   
\end{align*} and  \[g^*\mleft(\delta\mright) =  c_1 \mleft(q+\delta\mright) \mleft(1-q - \delta\mright) + c_2 \mleft( q\mleft(q+\delta\mright) + \mleft(1-q\mright) \mleft(1-q-\delta\mright)  \mright) + c_4 \mleft(1-q\mright) q,\] 
such that \[
-\Lambda_{X}^*\mleft(0\mright)  + \Lambda_{X'}^*\mleft(0\mright) = \log \mleft(2\sqrt{f^*\mleft(\delta\mright)}+c_0\mright) - m \log\mleft(2\sqrt{g^*\mleft(\delta\mright)}+c_0\mright),\] and we again have to show \eqref{Cramer_Final}, i.e. \[
f^{*'}\mleft(\delta\mright) \geq m  g^{*'}\mleft(\delta\mright)  \frac{\sqrt{f^*\mleft(\delta\mright)}\mleft(2\sqrt{f^*\mleft(\delta\mright)}+c_0\mright)}{\sqrt{g^*\mleft(\delta\mright)}\mleft(2\sqrt{g^*\mleft(\delta\mright)}+c_0\mright)} \] as we already know $-\Lambda_{X}^*\mleft(0\mright)  + \Lambda_{X'}^*\mleft(0\mright)$ to be positive for $\delta=0$.  This time,
\begin{align*}
g^{*'}\mleft(\delta\mright) &= c_1 \mleft(1-2\mleft(q+\delta\mright)\mright) - c_2 \mleft(1-2q\mright)
\end{align*} and 
\begin{align*}
f^{*'}\mleft(\delta\mright) &= c_1 \mleft(1-2M_m\mleft(q+\delta\mright)\mright) m  \binom{m-1}{\frac{m-1}{2}} \mleft(q+\delta\mright)^\frac{m-1}{2} \mleft(1-q-\delta\mright)^\frac{m-1}{2} 
 \\ &+
c_2 M_m\mleft(q\mright) m  \binom{m-1}{\frac{m-1}{2}} \mleft(q+\delta\mright)^\frac{m-1}{2} \mleft(1-q+\delta\mright)^\frac{m-1}{2} 
\\ &- 
c_2 \mleft(1-M_m\mleft(q\mright)\mright)   m  \binom{m-1}{\frac{m-1}{2}} \mleft(q+\delta\mright)^\frac{m-1}{2} \mleft(1- q+\delta\mright)^\frac{m-1}{2}
\\ & = 
\mleft(c_1 \mleft(1-2M_m\mleft(q+\delta\mright)\mright) - c_2 \mleft(1-2M_m\mleft(q\mright)\mright)\mright)  m  \binom{m-1}{\frac{m-1}{2}} \mleft(q+\delta\mright)^\frac{m-1}{2} \mleft(1- q-\delta\mright)^\frac{m-1}{2}.
\end{align*}
We note that \[c_1-c_2 = \mleft(\mleft(1-p_w\mright)p_b^0\mright)^2 - \mleft(1-p_w\mright)p_b^0 p_w \mleft(1-p_b^1\mright),\] which is positive if  \[\mleft(1-p_w\mright)p_b^0 -  p_w \mleft(1-p_b^1\mright)>0,\] i.e. \[\mleft(1-p_w\mright)p_b^0 + p_w \mleft(p_b^1-1\mright)>0,\] which we assumed to be true. Correspondingly, $c_1>c_2$ and because $1-2\mleft(q+\delta\mright)$ and $1-2M_m\mleft(q+\delta\mright)$ are monotonically falling in $\delta$, both $f^{*'}$ and $g^{*'}$ are negative. As such, \eqref{Cramer_Final} reduces to 
\begin{align*}\frac{f^{*'}}{mg^{*'}} & = 
 \binom{m-1}{\frac{m-1}{2}} \mleft(q+\delta\mright)^\frac{m-1}{2} \mleft(1- q-\delta\mright)^\frac{m-1}{2}  \frac{\mleft(c_1 \mleft(1-2M_m\mleft(q+\delta\mright)\mright) - c_2 \mleft(1-2M_m\mleft(q\mright)\mright)\mright)     }{c_1 \mleft(1-2\mleft(q+\delta\mright)\mright) - c_2 \mleft(1-2q\mright)} \\ &  \leq     \frac{\sqrt{f^*\mleft(\delta\mright)}\mleft(2\sqrt{f^*\mleft(\delta\mright)}+c_0\mright)}{\sqrt{g^*\mleft(\delta\mright)}\mleft(2\sqrt{g^*\mleft(\delta\mright)}+c_0\mright)}.
\end{align*} 
To get a better handle on this inequality, we need the following lemma:
\begin{lemma}\label{lemma:magic}
    Let $c_1,c_2$ be positive and $A,B,C,D$ be negative constants such that $c_1 A-c_2 B < 0$ and $c_1 C-c_2 D < 0$.
    Then $\frac{c_1 A-c_2 B}{c_1 C -c_2 D}\leq \frac{A}{C}$ is true if and only if $CB\geq DA$.
\end{lemma}
\begin{proof}
    \begin{align*}
        \frac{c_1 A-c_2 B}{c_1 C -c_2 D}\leq \frac{A}{C}  & \iff
         c_1 A-c_2 B\geq \frac{A\mleft(c_1 C -c_2 D\mright)}{C}  \\& \iff 
          C\mleft(c_1 A-c_2 B\mright)\leq A\mleft(c_1 C -c_2 D\mright)  \\& \iff 
          c_1 C A - c_2 C  B\leq c_1 C A -c_2 D A  \\& \iff 
           - c_2 C  B\leq  -c_2 D A  \\& \iff 
            C  B\geq   D A 
    \end{align*}
\end{proof}
We set \[A=\mleft(1-2M_m\mleft(q+\delta\mright)\mright),\] \[B=\mleft(1-2M_m\mleft(q\mright)\mright),\] \[C=\mleft(1-2\mleft(q+\delta\mright)\mright),\] \[D=\mleft(1-2q\mright),\] such that $CB\geq DA$ is equivalent to \[\mleft(1-2\mleft(q+\delta\mright)\mright)\mleft(1-2M_m\mleft(q\mright)\mright)\geq \mleft(1-2q\mright) \mleft(1-2M_m\mleft(q+\delta\mright)\mright),\] or \[\mleft(1-2M_m\mleft(q\mright)\mright)\leq \mleft(1-2q\mright) \frac{\mleft(1-2M_m\mleft(q+\delta\mright)\mright)}{\mleft(1-2\mleft(q+\delta\mright)\mright)},\] i.e.  \[\frac{\mleft(1-2M_m\mleft(q\mright)\mright)}{\mleft(1-2q\mright)}\geq  \frac{\mleft(1-2M_m\mleft(q+\delta\mright)\mright)}{\mleft(1-2\mleft(q+\delta\mright)\mright)},\] which is equivalent to 
\[1+ 2\sigma\mleft(m,q\mright)\geq  1+ 2\sigma\mleft(m,q+\delta\mright),\] which holds as $\sigma\mleft(m,x\mright)$ is clearly monotonically decreasing in $x$ for $x>0.5$. 
 
Lemma \ref{lemma:magic} allows us to upper bound 
\begin{align*}
     \frac{\mleft(c_1 \mleft(1-2M_m\mleft(q+\delta\mright)\mright) - c_2 \mleft(1-2M_m\mleft(q\mright)\mright)\mright)     }{c_1 \mleft(1-2\mleft(q+\delta\mright)\mright) - c_2 \mleft(1-2q\mright)} & \leq
      \frac{\mleft(1-2M_m\mleft(q+\delta\mright)\mright)}{1-2\mleft(q+\delta\mright)}  \\ &= 
       1 + 2 \sigma\mleft(m,q+\delta\mright). 
\end{align*} Correspondingly, \eqref{Cramer_Final} reduces to 
\begin{align}\label{delta_left}&
 \binom{m-1}{\frac{m-1}{2}} \mleft(q+\delta\mright)^\frac{m-1}{2} \mleft(1- q-\delta\mright)^\frac{m-1}{2} \mleft(1+2\sigma\mleft(m,q+\delta\mright)\mright) \nonumber\\& \leq     \frac{\sqrt{f^*\mleft(\delta\mright)}\mleft(2\sqrt{f^*\mleft(\delta\mright)}+c_0\mright)}{\sqrt{g^*\mleft(\delta\mright)}\mleft(2\sqrt{g^*\mleft(\delta\mright)}+c_0\mright)}. 
\end{align}
To control this, we need another lemma: 
\begin{lemma}\label{lemma:mixed_fraction}
    Let $c,f_1,f_2,g_1,g_2>0$; $f_1 \leq g_1$ and $f_2\geq g_2$. Then 
    \[\frac{f_1}{g_1}\leq\frac{2\mleft(f_1+f_2\mright)+c \sqrt{f_1+f_2}}{2\mleft(g_1+g_2\mright)+c \sqrt{g_1+g_2}}\]
\end{lemma}
\begin{proof}
    We first note that \[
    \mleft(f_1-g_1\mright)f_1 g_1 \leq g_1^2 f_2 - f_1^2 g_2,
    \] as the left side is always negative because $f_1\leq g_1$, while the right side is always positive as $g_1\geq f_1$ and $f_2\geq g_2$. With this, we calculate
    \begin{align*}
        & \mleft(f_1-g_1\mright)f_1 g_1 \leq g_1^2 f_2 - f_1^2 g_2 \\ & \iff
        f_1^2 g_1 + f_1^2 g_2 \leq g_1^2 f_1 +g_1^2 f_2 \\
        & \iff
        f_1^2 \mleft(g_1 + g_2\mright) \leq g_1^2 \mleft(f_1 + f_2\mright) \\
        & \iff
        f_1 \sqrt{g_1 + g_2} \leq g_1 \sqrt{f_1 + f_2}.
    \end{align*}
    With this, 
    \begin{align*}
        & \frac{f_1}{g_1}\leq\frac{2\mleft(f_1+f_2\mright)+c \sqrt{f_1+f_2}}{2\mleft(g_1+g_2\mright)+c \sqrt{g_1+g_2}} \\& \iff
        f_1 \mleft(2\mleft(g_1+g_2\mright)+c \sqrt{g_1+g_2}\mright) \leq g_1 \mleft(2\mleft(f_1+f_2\mright)+c \sqrt{f_1+f_2}\mright) \\& \iff
         f_1 \mleft(2g_2+c \sqrt{g_1+g_2}\mright) \leq g_1 \mleft(2f_2+c \sqrt{f_1+f_2}\mright).
    \end{align*}
    The inequality now holds for the second terms on each side by our previous calculations, and for the first terms on each side as $f_1\leq g_1$ and $g_2 \leq f_2$.
\end{proof}
We set \[f_1 =  c_1 M_m\mleft(q+\delta\mright) \mleft(1-M_m\mleft(q+\delta\mright)\mright) + c_4 \mleft(1-M_m\mleft(q\mright)\mright)M_m\mleft(q\mright),\]
\[g_1 =  c_1 \mleft(q+\delta\mright) \mleft(1-q - \delta\mright) + c_4 \mleft(1-q\mright) q ,\]
as well as 
\[f_2 =  c_2\mleft(M_m\mleft(q\mright)M_m\mleft(q+\delta\mright) +\mleft(1-M_m\mleft(q\mright)\mright)\mleft(1-M_m\mleft(q+\delta\mright)\mright) \mright) ,\] and 
\[g_2 =  c_2 \mleft( q\mleft(q+\delta\mright) + \mleft(1-q\mright) \mleft(1-q-\delta\mright)  \mright) ,\]
such that \[f_1+f_2 = f^*\mleft(\delta\mright)\] and \[g_1+g_2 = g^*\mleft(\delta\mright).\]
If we can prove the preconditions for \ref{lemma:mixed_fraction}, \eqref{delta_left} will reduce to \begin{equation}\label{delta_right}
 \binom{m-1}{\frac{m-1}{2}} \mleft(q+\delta\mright)^\frac{m-1}{2} \mleft(1- q-\delta\mright)^\frac{m-1}{2} \mleft(1+2\sigma\mleft(m,q+\delta\mright)\mright) \leq     \frac{f_1}{g_1}.
\end{equation}
$f_1\leq g_1$ is easy to see, based on $M_m\mleft(q\mright)$ increasing in $m$, and $x\mleft(1-x\mright)$ decreasing in $x$ for $x>0.5.$ We can thus focus on showing $g_2 \leq f_2$, i.e. \begin{align} \label{c2terms}
& \mleft( q\mleft(q+\delta\mright) + \mleft(1-q\mright) \mleft(1-q-\delta\mright)  \mright)  \\ & \leq \mleft(M_m\mleft(q\mright)M_m\mleft(q+\delta\mright) +\mleft(1-M_m\mleft(q\mright)\mright)\mleft(1-M_m\mleft(q+\delta\mright)\mright) \mright). \nonumber
\end{align}
At $\delta=1-q$, \eqref{c2terms} becomes \[q \leq  M_m\mleft(q\mright), 
\] which is clearly true. At $\delta=0$, we get \[q^2 + \mleft(1-q\mright)^2 \leq  M_m\mleft(q\mright)^2 + \mleft(1- M_m\mleft(q\mright)\mright)^2. \] We note that \[x^2 + \mleft(1-x\mright)^2 = 1 + 2 \mleft(x^2 - x\mright)  \] has the derivative $4x - 2$, which is positive for $x>0.5$. Correspondingly, the $M_m\mleft(q\mright)$ term is larger than the $q$ term. Having shown that \eqref{c2terms} holds at both extreme values for $\delta$, it is sufficient for Lemma \ref{lemma:mixed_fraction} to hold to show that the second derivative of 
\begin{align*} & \mleft( q\mleft(q+\delta\mright) + \mleft(1-q\mright) \mleft(1-q-\delta\mright)  \mright)   - \mleft(M_m\mleft(q\mright)M_m\mleft(q+\delta\mright) +\mleft(1-M_m\mleft(q\mright)\mright)\mleft(1-M_m\mleft(q+\delta\mright)\mright) \mright)\end{align*} with respect to $\delta$ is positive, such that the function is convex. As the left term is linear in $\delta$, this derivative equals \[ - M_m\mleft(q\mright) \frac{d^2}{d^2 \delta} M_m\mleft(q+\delta\mright)  + \mleft(1- M_m\mleft(q\mright)\mright)  \frac{d^2}{d^2 \delta} M_m\mleft(q+\delta\mright),\] which equals \[ \mleft(1- 2 M_m\mleft(q\mright)\mright)  \frac{d^2}{d^2 \delta} M_m\mleft(q+\delta\mright)\] and thus has the opposite sign of $\frac{d^2}{d^2 \delta} M_m\mleft(q+\delta\mright)$, which is negative due to the well-known concavity of the majority vote in $M_m\mleft(x\mright)$ in $x$ for $x>0.5$ \citep{boland1989modelling}. 

To prove \eqref{delta_right}, we need one last lemma:
\begin{lemma}\label{lemma:mono_frac}
    Let $A,B,C,D>0$ and $AD\leq BC$. Then, $\frac{A}{C}\leq\frac{A+B}{C+D}$
\end{lemma}
\begin{proof}
    \begin{align*}
        \frac{A}{C}\leq\frac{A+B}{C+D}  \iff
        AC + AD \leq AC + BC \iff
        AD \leq BC 
    \end{align*}
\end{proof}
We set \[A = c_1 M_m\mleft(q+\delta\mright) \mleft(1-M_m\mleft(q+\delta\mright)\mright),\] \[B = c_4 \mleft(1-M_m\mleft(q\mright)\mright)M_m\mleft(q\mright),\] \[C = c_1 \mleft(q+\delta\mright) \mleft(1-q-\delta\mright),\] \[D = c_4 \mleft(1-q\mright)q,\] such that \[f_1 = A+B\] and \[g_1 = C+D.\] If we can show that $AD\leq BC$,  \eqref{delta_right} would reduce to  \begin{align}\label{delta_final}
 & \binom{m-1}{\frac{m-1}{2}} \mleft(q+\delta\mright)^\frac{m-1}{2} \mleft(1- q-\delta\mright)^\frac{m-1}{2} \mleft(1+2\sigma\mleft(m,q+\delta\mright)\mright) \nonumber\\& \leq     \frac{M_m\mleft(q+\delta\mright) \mleft(1-M_m\mleft(q+\delta\mright)\mright)}{\mleft(q+\delta\mright) \mleft(1-q-\delta\mright)}, 
\end{align}
which is equivalent to \eqref{prestirling3} and true by the calculations in section \ref{sec:ld}.
 $AD\leq BC$ is equivalent to \[
M_m\mleft(q+\delta\mright) \mleft(1-M_m\mleft(q+\delta\mright)\mright) \mleft(1-q\mright)q \leq \mleft(1-M_m\mleft(q\mright)\mright)M_m\mleft(q\mright) \mleft(q+\delta\mright) \mleft(1-q-\delta\mright).\] This is again clearly true for $\delta=0$ where both sides are equal, such that it is sufficient to show that 
 \[ \frac{M_m\mleft(q+\delta\mright) \mleft(1-M_m\mleft(q+\delta\mright)\mright) \mleft(1-q\mright)q }{\mleft(1-M_m\mleft(q\mright)\mright)M_m\mleft(q\mright) \mleft(q+\delta\mright) \mleft(1-q-\delta\mright)}\] or \[\frac{\mleft(1-q\mright)q}{\mleft(1-M_m\mleft(q\mright)\mright)M_m\mleft(q\mright)}\frac{M_m\mleft(q+\delta\mright) \mleft(1-M_m\mleft(q+\delta\mright)\mright)}{\mleft(q+\delta\mright) \mleft(1-q-\delta\mright)}\] is maximized at $\delta=0$. As the first term does not depend on $\delta$, we only need to analyze the second term. Reparameterizing $x=q+\delta$, it is thus sufficient to show that \[\frac{M_m\mleft(x\mright)\mleft(1-M_m\mleft(x\mright)\mright)}{x\mleft(1-x\mright)} \] decreases monotonically in $x$. We take derivatives with respect to $x$, obtaining \[
\frac{\mleft(1-2M_m\mleft(x\mright)\mright) m \binom{m-1}{\frac{m-1}{2}} x^\frac{m+1}{2} \mleft(1-x\mright)^\frac{m+1}{2} - M_m\mleft(x\mright)\mleft(1-M_m\mleft(x\mright)\mright) \mleft(1-2x\mright) }{x^2\mleft(1-x\mright)^2}.
 \]
 This is negative, whenever \[ \mleft(1-2M_m\mleft(x\mright)\mright) m \binom{m-1}{\frac{m-1}{2}} x^\frac{m+1}{2} \mleft(1-x\mright)^\frac{m+1}{2} \leq  M_m\mleft(x\mright)\mleft(1-M_m\mleft(x\mright)\mright) \mleft(1-2x\mright)\] or equivalently \[ \frac{1-2M_m\mleft(x\mright)}{1-2x} m \binom{m-1}{\frac{m-1}{2}} x^\frac{m+1}{2} \mleft(1-x\mright)^\frac{m+1}{2} \geq  M_m\mleft(x\mright)\mleft(1-M_m\mleft(x\mright)\mright),\] i.e.   \[ \mleft(1+2\sigma\mleft(m,x\mright)\mright) m \binom{m-1}{\frac{m-1}{2}} x^\frac{m+1}{2} \mleft(1-x\mright)^\frac{m+1}{2} \geq  M_m\mleft(x\mright)\mleft(1-M_m\mleft(x\mright)\mright).\] As both sides tend to zero for $x\rightarrow 1$, it is sufficient to show that the right term increases more slowly as $x$ decreases, i.e.  \begin{equation}\label{derivative_reverse} \frac{d}{dq}\mleft(\mleft(1+2\sigma\mleft(m,x\mright)\mright) m \binom{m-1}{\frac{m-1}{2}} x^\frac{m+1}{2} \mleft(1-x\mright)^\frac{m+1}{2}\mright) \leq \frac{d}{dq}\mleft(M_m\mleft(x\mright)\mleft(1-M_m\mleft(x\mright)\mright)\mright).\end{equation} Note, that this equation is the reverse of \eqref{derivative_base}, but with an additional factor of $m$ on the left side. Repeating the calculations from Section \ref{sec:ld}, \eqref{derivative_reverse} reduces to \[ 
m\mleft(\frac{m+1}{2} + \frac{q\mleft(1-q\mright)}{1-2M_m\mleft(q\mright)} 2\frac{d}{dq} \sigma\mleft(m,q\mright)\mright)\geq m 
\]  or 
\[ 
\frac{m+1}{2} + \frac{q\mleft(1-q\mright)}{1-2M_m\mleft(q\mright)} 2\frac{d}{dq} \sigma\mleft(m,q\mright)\geq 1.
\]
The $\frac{q\mleft(1-q\mright)}{1-2M_m\mleft(q\mright)} 2\frac{d}{dq} \sigma\mleft(m,q\mright)$ term is positive, as both the first and the second factor are clearly negative, such that the equation holds, finishing our proof of \[\Pr\mleft(\sum^n_i G_i\mleft(M_m\mleft(q\mright),p\mright)>0\mright)<\Pr\mleft(\sum^{mn}_i G_i\mleft(q,p\mright)>0\mright).\]
It remains to show that for fixed $q_b\geq q_w$ and $m>1$ uneven, $G$ in the heterogeneous case stochastically dominates $G$ for the homogeneous whenever assumption \ref{ass:jensen} holds. This would imply that the sum of $G_i$ follows the same dominance relation, such that the probability of correctly identifying $c_b$ is larger for the $m-$label case assuming homogeneity rather than explicitly modelling heterogeneity. We note that $\Pr\mleft(G\mleft(q,p\mright)=0\mright)$ does not depend on $q$, such that it is sufficient to show that $\Pr\mleft(G\mleft(q,p\mright)=1\mright)$ is larger in the homogeneous case. We rewrite 
\begin{align*}
    & \frac{\mleft(1-p_w\mright)p_b^0}{p_w \mleft(1-p_b^1\mright)}\mleft(M_m\mleft(q_b\mright)-\E_{x}[M_m\mleft(q\mleft(x\mright)\mright)|E_b]\mright) \geq   M_m\mleft(q_w\mright)-\E_{x}[M_m\mleft(q\mleft(x\mright)\mright)|E_w] \\  \iff &
    \mleft(1-p_w\mright)p_b^0\mleft(M_m\mleft(q_b\mright)-\E_{x}[M_m\mleft(q\mleft(x\mright)\mright)|E_b]\mright) \\ & \geq  p_w \mleft(1-p_b^1\mright) \mleft(M_m\mleft(q_w\mright)-\E_{x}[M_m\mleft(q\mleft(x\mright)\mright)|E_w]\mright) \\
      \iff &
      \mleft(1-p_w\mright)p_b^0 M_m\mleft(q_b\mright)  -  p_w \mleft(1-p_b^1\mright) M_m\mleft(q_w\mright)  \\ & \geq \mleft(1-p_w\mright)p_b^0 \E_{x}[M_m\mleft(q\mleft(x\mright)\mright)|E_b]  - p_w \mleft(1-p_b^1\mright) \E_{x}[M_m\mleft(q\mleft(x\mright)\mright)|E_w]\\
      \iff & 
        \mleft(1-p_w\mright)p_b^0 M_m\mleft(q_b\mright)  -  p_w \mleft(1-p_b^1\mright) \mleft(1-M_m\mleft(q_w\mright)\mright)  \\ & \geq \mleft(1-p_w\mright)p_b^0 \E_{x}[M_m\mleft(q\mleft(x\mright)\mright)|E_b]  - p_w \mleft(1-p_b^1\mright) \mleft(1-\E_{x}[M_m\mleft(q\mleft(x\mright)\mright)|E_w]\mright) \\ 
        \iff &  \Pr\mleft(G\mleft(M_m\mleft(q_b\mright),M_m\mleft(q_w\mright),p\mright)=1\mright) \\ & \geq \Pr\mleft(G\mleft(\E_{x}[M_m\mleft(q\mleft(x\mright)\mright)|E_b],\E_{x}[M_m\mleft(q\mleft(x\mright)\mright)|E_w],p\mright)=1\mright),
\end{align*} showing that the heterogeneous case is dominated by the homogeneous case under assumption \ref{ass:jensen}. 

\end{document}